\documentclass[10pt,twocolumn,letterpaper]{article}

\usepackage[pagenumbers]{cvpr} 

\def\maketitleappendix
   {
   \newpage
       \twocolumn[
        \centering
        \Large
        \vspace{0.5em}\textbf{Appendix} \\
        \vspace{1.0em}
       ] 
   }

\usepackage{times}
\usepackage{url}
\usepackage{epsfig}
\usepackage{graphicx}
\usepackage{amsmath}
\usepackage{amssymb}
\usepackage{multirow}
\usepackage{float}
\usepackage{enumitem}
\usepackage{pifont}
\usepackage{booktabs}
\usepackage{wasysym}
\usepackage{caption}
\usepackage{subcaption}
\usepackage{colortbl}
\usepackage{stfloats}
\usepackage{graphicx}
\usepackage{cuted}
\usepackage[accsupp]{axessibility}
\usepackage[dvipsnames]{xcolor}

\definecolor{lg}{HTML}{57cc99}  
\definecolor{lo}{HTML}{dc2f02} 

\newcommand{\cmark}{\textcolor{lg}{\ding{51}}}
\newcommand{\xmark}{\textcolor{lo}{\ding{55}}}
\newcommand{\smark}[1][orange]{\textcolor{#1}{\textbf{$\sim$}}}
\newcommand{\sword}{\text{\textdagger}}

\newcommand\mypar[1]{\par\vspace{1.0mm}\noindent\textbf{#1}\;\;}
\def\ourmodel{Vid2Sim\xspace}

\definecolor{cvprblue}{rgb}{0.21,0.49,0.74}
\usepackage[pagebackref,breaklinks,colorlinks,allcolors=cvprblue]{hyperref}

\def\paperID{} 
\def\confName{}
\def\confYear{}

\title{\ourmodel: Realistic and Interactive Simulation from Video for Urban Navigation}

\author{Ziyang Xie$^{1, 2}$ \quad Zhizheng Liu$^2$ \quad Zhenghao Peng$^2$ \quad Wayne Wu$^2$ \quad Bolei Zhou$^2$ \\
$^1$University of Illinois Urbana-Champaign \quad $^2$University of California, Los Angeles 
\\[0.5em]
\url{https://metadriverse.github.io/vid2sim/}
}

\begin{document}
\twocolumn[{%
    \renewcommand\twocolumn[1][]{#1}%
    \maketitle
    \vspace{-15pt}
    \includegraphics[width=0.97\linewidth]{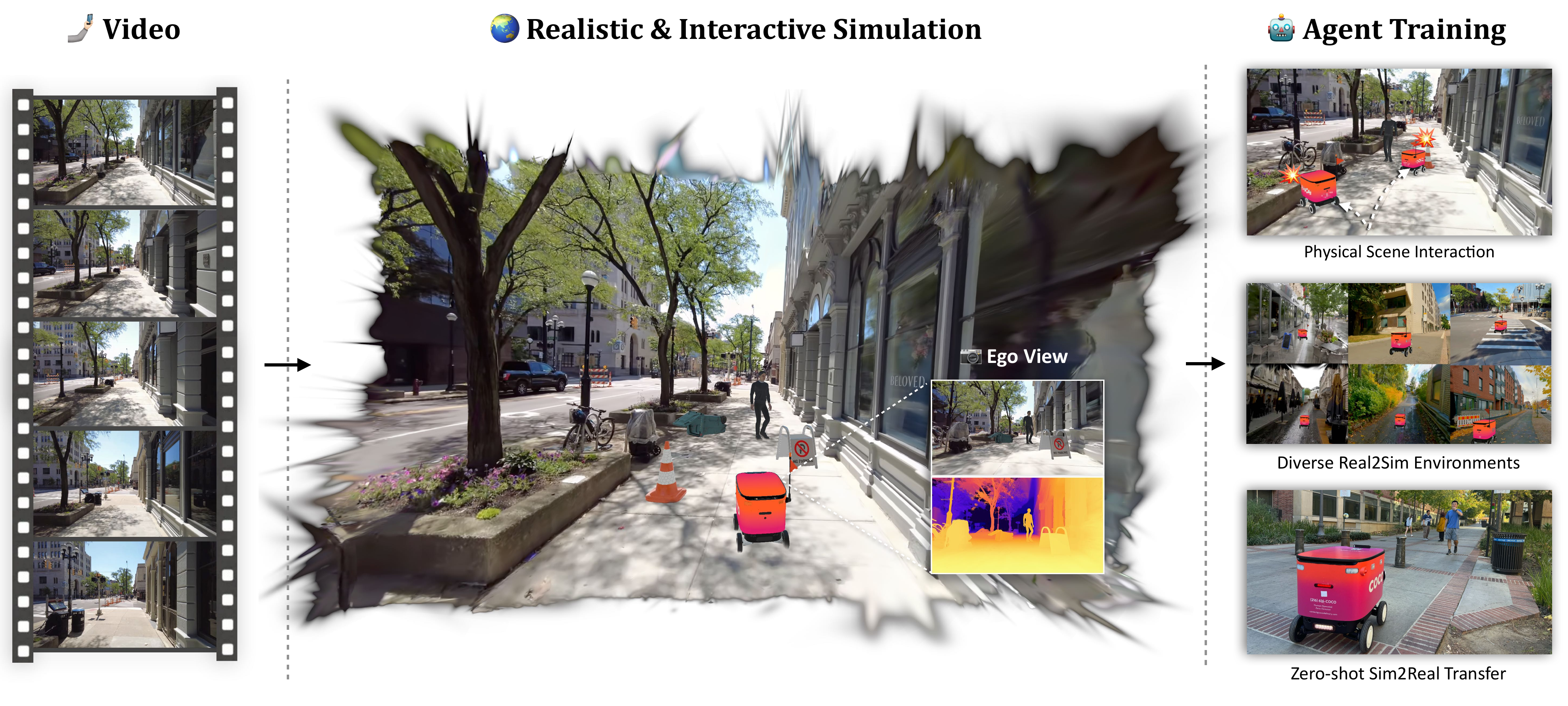}
    \vspace{-8pt}
    \captionof{figure}{\ourmodel converts monocular video captured by a hand-held camera into realistic and interactive 3D simulation environments. It facilitates RL training of navigation agents in digital twins of urban scenes and provides realistic observations like RGB and depth to reduce the sim-to-real gap. The pink mobile robot in the image is a food delivery bot that avoids collisions with pedestrians and obstacles. 
    }
    \label{fig:teaser}
    \vspace{8pt}
}]

\begin{abstract}
Sim-to-real gap has long posed a significant challenge for robot learning in simulation, preventing the deployment of learned models in the real world. Previous work has primarily focused on domain randomization and system identification to mitigate this gap. However, these methods are often limited by the inherent constraints of the simulation and graphics engines. 
In this work, we propose \ourmodel, a novel framework that effectively bridges the sim2real gap through a scalable and cost-efficient real2sim pipeline for neural 3D scene reconstruction and simulation. Given a monocular video as input, \ourmodel can generate photo-realistic and physically interactable 3D simulation environments to enable the reinforcement learning of visual navigation agents in complex urban environments. 
Extensive experiments demonstrate that \ourmodel significantly improves the performance of urban navigation in the digital twins and real world by 31.2\% and 68.3\% in success rate compared with agents trained with prior simulation methods. 
\end{abstract}    
\vspace{-2mm}
\section{Introduction}
\label{sec:intro}


Developing intelligent agents navigating and interacting within complex urban environments is crucial for many robotic applications like food delivery bots and assistive electric wheelchairs. Real-world experimentation is often challenging due to safety risks and efficiency constraints. In recent years, learning in simulation~\cite{makoviychuk2021isaacgymhighperformance, todorov2012mujoco, habitat19iccv} has become an essential tool to provide a safe, controlled, and cost-effective alternative for training agents on complex tasks such as robotic manipulation~\cite{liqian2023sim2real2, pmlr-v87-kalashnikov18a, li24simpler} and autonomous driving~\cite{alexey2017carla, yang2023unisim, li2021metadrive, wu2024metaurban}. However, transferring the models trained in a simulated environment into the real world remains challenging due to the significant sim-to-real gap. 


Substantial efforts have been made to bridge this gap. Prior approaches often leverage domain randomization~\cite{domainRandom2017, Sadeghi2017CAD2RL, Tremblay2018Training} and system identification~\cite{Chebotar2019Closing, Tan2018SimToReal, Peng2018SimToReal} methods that enhance the agent's robustness by simulating real-world noises and aligning agent dynamic model with the real-world settings. While these methods have achieved some success, their effectiveness is fundamentally limited by the simulator's capabilities. Traditional simulators often fail to provide realistic observations, dynamic interactions, and diverse environmental variations. It is thus difficult to accurately reproduce the digital twins of real-world scenarios for agent training, limiting the agent's capacity to generalize effectively beyond the simulation environments. Recent advances in neural rendering techniques such as NeRF~\cite{mildenhall2020nerf} and 3DGS~\cite{ kerbl3Dgaussians} emerge as a potential solution to bridge the sim2real gap by reconstructing realistic 3D scenes from real-world data. However, most works~\cite{mildenhall2020nerf, kerbl3Dgaussians, barron2022mipnerf360, Huang2DGS2024} only focus on enhancing photorealism for novel view synthesis and often fall short in constructing fully interactive environments that can support embodied agent training. 
While recent work Video2Game~\cite{xia2024video2game} attempts to extend NeRF-based neural reconstructions for interactive game development, its applications remain limited to gaming contexts, and its visual fidelity is limited by the textured mesh representation. Consequently, its effectiveness in training generalizable embodied agents remains constrained.



To tackle these issues, in this work, we present \ourmodel, a novel real-to-sim (real2sim) framework that can convert causal videos captured in the real world into a fully realistic and physical interactive simulation environment and facilitate the reinforcement learning of urban navigation agents with minimal sim-to-real gap. By leveraging abundant web video data, our method offers a scalable and cost-efficient solution to build a realistic and diverse simulation environment for embodied agent training, as illustrated in Figure \ref{fig:teaser}. Our pipeline consists of two main components: 1) geometry-consistent scene reconstruction and 2) realistic interactive simulation construction.
Given a video as input, we design a geometry-consistent scene reconstruction approach that utilizes monocular cues to regularize Gaussian Splatting training in a scale-invariant way, thereby enhancing the reconstruction of fine-grained geometric details. In addition, we introduce a screen-space 2D covariance culling method to improve post-training rendering quality when the agent's camera trajectory deviates substantially from the training views.

To achieve realistic and interactive scene reconstruction, we propose a novel hybrid scene representation that combines Gaussian Splatting (GS) representation with mesh primitives. The GS provides real-time, photorealistic visuals for the agent, while the underlying mesh enables physical interactions and collision detection, which are essential for navigation training. Using \ourmodel, we generate a diverse dataset of real2sim environments from web videos, encompassing a wide range of real-world urban settings for navigation agent training. These environments are further populated with static obstacles and other dynamic road users to replicate challenging real-world navigation scenarios. We augment these scenes through 3D scene editing and particle systems in different lighting, styles, and weather conditions to further improve the generalizability of our agents.
Experiments in both simulation and real-world settings show that agents trained with \ourmodel achieve substantially higher success rates and exhibit zero-shot sim2real transfer ability in real-world deployment. Compared to the agents trained with traditional simulation methods, \ourmodel agent demonstrates enhanced robustness and lower collision rate, highlighting its potential as a scalable and efficient solution for bridging the sim-to-real gap. We summarize our contributions as follows:

\begin{enumerate} 
    \item \textbf{Real2Sim simulation environment from monocular video}: We introduce \ourmodel, a novel real-to-sim (real2sim) framework that can convert monocular videos into photorealistic and physically interactive simulation environments. 

    \item \textbf{Geometry-consistent reconstruction with hybrid representation}: We develop a scene reconstruction method that improves scene reconstruction quality and agent visual observation quality through geometry-consistent GS training and screen-space covariance culling.
   We further propose a hybrid scene representation that combines GS with mesh primitives to enable photorealistic rendering and accurate physical interactions for urban navigation training.

    \item \textbf{Comprehensive and diverse scene augmentation}: We present a scene composition method that integrates static obstacles and dynamic agents to construct interactive and diverse navigation environments, replicating complex real-world navigation challenges. Our method also supports extensive scene augmentation to environment layouts, lighting conditions, and weather for robust visual navigation training in complex urban environments. 

\end{enumerate}


\section{Related Work}
\label{sec:formatting}

\mypar{3D scene reconstruction}
3D scene reconstruction has long been an active research topic in computer vision that aims to reconstruct the 3D scene representation from 2D image inputs. Recently, advances in neural reconstruction methods have achieved impressive results. Techniques like Neural Radiance Fields (NeRF)~\cite{mildenhall2020nerf} and Gaussian Splatting (3DGS)~\cite{kerbl3Dgaussians} have demonstrated impressive abilities to reconstruct complex scenes from 2D inputs with photorealistic rendering. Building upon these methods, various extensions have been developed to address specific challenges, such as few-shot 3D reconstruction~\cite{Yang2023FreeNeRF, xu2024fewshotnerfadaptiverendering, Jain_2021_ICCV, Niemeyer2021Regnerf, zhu2023FSGS, chung2023depth} and reconstruction in unbounded, in-the-wild settings~\cite{martinbrualla2020nerfw, zhang2024gaussian, barron2022mipnerf360}, further enhancing their applicability to real-world data. Other works~\cite{wang2021neus, li2023neuralangelo, Huang2DGS2024, guedon2023sugar, wolf2024gs2mesh, Yu2024GOF, Wu2024gsrec} have focused on surface reconstruction task to better reconstruct object surface and capture the scene geometry. Despite these advancements, most works remain limited to photo-realistic reconstructions and cannot simulate physical interactions. In this work, we explore the possibility of extending GS representation to create a physically interactive and realistic simulation from monocular videos, facilitating agent training in real-world consistent digital twins.

\mypar{Sim-to-Real transfer}
The sim-to-real (sim2real) gap remains a significant challenge in deploying AI models from simulation to real-world environments. Traditional approaches like domain randomization \cite{domainRandom2017, Sadeghi2017CAD2RL, Tremblay2018Training} and system identification \cite{Chebotar2019Closing, Tan2018SimToReal, Peng2018SimToReal} aim to bridge this gap by mimicking real-world variations and aligning simulations with actual setups. However, their effectiveness is limited by the inherent fidelity of simulation environments. 

Recent efforts \cite{wang2023drive, swerdlow2024streetview, li2023drivingdiffusion} have leveraged image generative models to produce high-quality 2D observations for agents training. These methods often lack realistic physics, hindering physical interactions and closed-loop evaluations necessary for embodied agent training. Other studies \cite{yu2024lucidsim, zhou2024simgen} attempt to integrate physics simulators with controlled 2D generation to ensure consistent observation simulation with physical interactions. However, they still rely on traditional simulators with limited environments for interaction and image layout control, preventing them from fully capturing real-world complexity. In our work, we resolve these issues by introducing a scalable and cost-efficient real2sim pipeline that can construct photo-realistic and interactive simulations from real-world videos. As a result, our method can significantly reduce the sim2real gap while avoiding the appearance and cost limitations introduced by traditional graphics simulators.

\mypar{Data-driven simulation}
Data-driven simulation generation is crucial for creating realistic, diverse environments for embodied agents training. Traditional systems~\cite{wu2024metaurban, li2021metadrive, habitat19iccv, szot2021habitat, xiazamirhe2018gibsonenv} use data-driven simulators to support navigation and manipulation tasks within diverse simulated environments. Despite progress, their scalability and fidelity remain limited. Recent works like Sim-on-Wheels~\cite{shen2023simonwheels} try to resolve this issue by incorporating a vehicle-in-the-loop framework that directly integrates real-world driving tests with virtual, safety-critical simulations. 
However, it still faces cost inefficiency and high time demands due to required real-world vehicle integration and continuous operation. Video2Game~\cite{xia2024video2game} leverages neural radiance fields to create interactive scenes from video for game engines, but its visual quality is constrained by textured mesh representation, and it mainly targets game development, limiting its use in embodied AI simulation. Our proposed method introduces a hybrid scene representation that combines GS with structured meshes, enabling the generation of photorealistic, physically interactive simulations from real-world monocular video input in a data-efficient way.
\section{Preliminary: 3D Gaussian Splatting}

3DGS~\cite{kerbl3Dgaussians} has emerged as a popular point-based method for 3D scene reconstruction that explicitly represents the scene as a set of 3D Gaussian primitives. For each gaussian splat $\mathcal{G}_i(\mathbf{x})$, it's been parametrized by its means $\mu_i \in \mathbb{R}^3$, 3D covariance $\Sigma_i \in \mathbb{R}^{3\times3}$, opacity $\mathbf{o}_i$ and color $\mathbf{c}_i$ as:
\begin{equation}
\label{eq:gaussian-splat}
    \mathcal{G}_i(\mathbf{x}) = \exp{(-\frac{1}{2}(\mathbf{x}-\mu_i)^T\Sigma_i^{-1}(\mathbf{x}-\mu_i))}.
\end{equation}
During rendering, these 3D Gaussian splats $\mathcal{G}_i$ are projected onto the image plane as 2D Gaussians $\mathcal{G}_i^{'}$. The projection process can be represented as: $\Sigma_{i}^{'} = JW\Sigma_{i}W^{T}J^{T}$, where $\Sigma^{'}_{i}$ represent the 2D screen space gaussians covariance, $W$ is the world-to-camera transformation matrix and $J$ is the Jacobian of the perspective projection equation. To maintain a positive semi-definite 3D covariance $\Sigma_i$ during optimization, $\Sigma_i$ is then reparametrized using a scaling matrix $S \in \mathbb{R}^{3}$ and a rotation matrix $R \in \mathbb{R}^{3\times3}$ as $\Sigma_i = R_iS_i S_i^{T} R_i^{T}$. The color of pixel $\mathbf{c}(x)$ can then be rendered through a volumetric alpha-blending process:
\begin{equation}
    \mathbf{c}(x) = \sum_{i \in N}{T_i\mathbf{c}_i\alpha_i(\mathbf{x})},\ \  T_i = \prod_{i = 1}^{i-1}(1-\alpha_i(\mathbf{x})),
\end{equation}
where $\alpha_i(\mathbf{x}) = \mathbf{o}_i\mathcal{G}_i(\mathbf{x})$ represents the alpha value of the Gaussian Splats $\mathcal{G}_i$ at point $\mathbf{x} \in \mathbb{R}^3$ and $\mathbf{c}_i$ is the color of $\mathcal{G}_i$ evaluated by its spherical harmonics (SH) coefficients. Similarly, we can extend and render out the per-pixel median depth and normal for the Gaussian Splats as: 
\begin{align}
\label{eq:render}
    \hat{\mathbf{D}}(x) &= \sum_{i \in N} T_i \mathbf{d}_i \alpha_i(\mathbf{x}),\ \ \hat{\mathbf{N}}(x) = \sum_{i \in N} \hat{\mathbf{n}}_i\alpha T_i,
\end{align}
where $\mathbf{d}_i$ is the $i^{\textit{th}}$ Gaussian Splat distance to the camera and $\hat{\mathbf{n}}_i$ is the normal direction based on the shortest axis direction of its covariance. Parameters of the 3DGS are then optimized by a photometric rendering loss in 2D image space.

While 3DGS effectively reconstructs visually realistic scenes, it struggles with accurate geometry reconstruction, tends to overfit training views, and cannot support physical interaction, limiting its use in interactive robotics learning. We introduce our \ourmodel framework in the next section to overcome these challenges.

\section{\ourmodel Framework}
\begin{figure*}[ht!]
    \centering
    \vspace{-2mm}
    \includegraphics[width=0.98\linewidth]{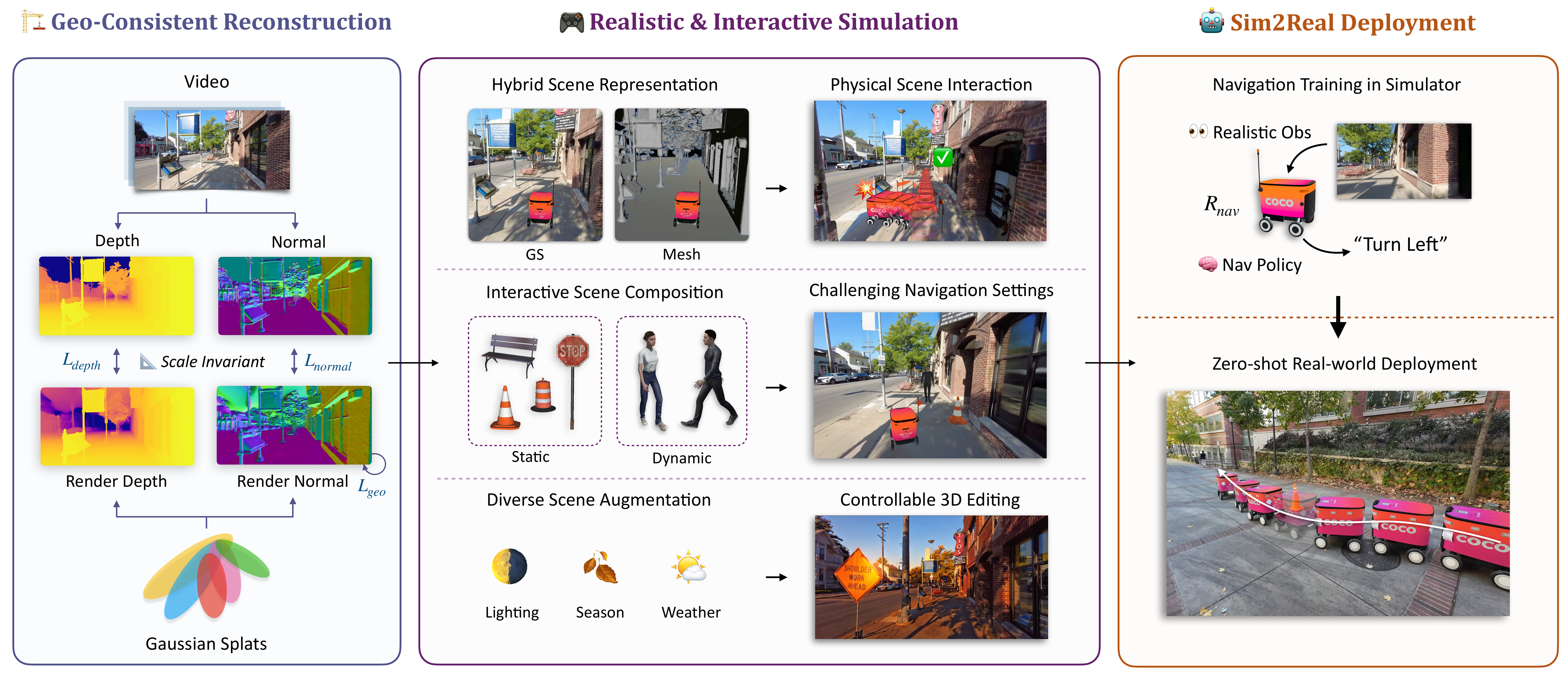}
     \caption{
     \ourmodel framework consists of three key stages: (1) Geometry-consistent reconstruction for high-quality environment creation, (2) building a realistic and interactive simulation with hybrid scene representation and diverse obstacle and scene augmentation for visual navigation training, and (3) Sim2Real validation through real-world deployment. 
     }
    \label{fig:model}
    \vspace{-4mm}
\end{figure*}

Given a monocular video, the goal of \ourmodel is to generate a realistic and physically interactive simulation environment for embodied navigation training with minimal sim-to-real gap. To achieve this, \ourmodel employs a two-stage pipeline: We first reconstruct a high-quality 3D scene representation with geometry-consistent Gaussian Splatting. In the second stage, we combine the reconstructed splats and mesh into a hybrid scene representation, building a photorealistic and interactive training environment with diverse obstacles and augmentations to support robust visual navigation training in complex environments.

\subsection{Geometry-Consistent Scene Reconstruction}
\label{sec:method-recon}

Accurate geometry reconstruction is important for agent navigation training to support accurate collision detection and realistic physical interactions. However, reconstructing the high-quality 3D structure of a scene from casual monocular video footage remains challenging due to the inherent geometric uncertainties and the lack of multi-view information.
To overcome this challenge, we propose a geometry-consistent reconstruction method that utilizes monocular cues to regularize GS training, enhancing geometry reconstruction for accurate agent-environment interactions.

\mypar{Scale-Invariant Geometry Supervision} Current advancements~\cite{depth_anything_v1, depth_anything_v2, birkl2023midas} in monocular depth estimation can be used to provide a strong geometry prior for in-the-wild scene reconstruction. However, most of these methods often implement a scale-shift-invariant (SSI) loss~\cite{birkl2023midas, miangolehSIDepth} to predict relative depth rather than absolute metric depth. Directly minimizing the differences between rendered and predicted depth may cause ambiguities, since the Gaussian Splats is initialized with a specific depth scale from Structure-from-Motion (SfM)~\cite{schoenberger2016sfm} point clouds,  

As illustrated in Fig.~\ref{fig:model}, we propose to use scale-invariant losses over the depth and normal to tackle that issue. Specifically, a patch-based normalized cross-correlation (NCC) loss is applied between the rendered depth $\hat{\mathbf{D}}$ and the predicted depth $\mathbf{D}$ generated from an off-the-shelf depth estimator~\cite{depth_anything_v2} for supervision. The patch-based NCC loss evaluates the local similarity between depth maps while being less sensitive to global scale discrepancies:

\begin{equation} 
\mathcal{L}_\mathrm{depth} = 1 - \frac{1}{\|\mathcal{P}\|}\sum_{p \in \mathcal{P}}\sum_{k=1}^{K^2} \frac{\hat{\mathbf{D}}_{p,k}^{'}\mathbf{D}_{p,k}^{'}}{\hat{\sigma}_p \sigma_p},
\end{equation}
where $\mathcal{P}$ is the set of all patches extracted from the depth map, and the inner sum over $k$ represents the summation of the NCC scores within a single patch of size $K \times K$. $\hat{\mathbf{D}}_{p,k}^{'}$ and $\mathbf{D}_{p,k}^{'}$ are the mean-centered values of the rendered and predicted depths at pixel $k$ within patch $p$, respectively. The $\hat{\sigma}_p$ and $\sigma_p$ are standard deviations of the rendered and predicted depth maps within the patch. This approach ensures that depth alignment is based on local structural similarity rather than absolute scale, which is more robust to noise and occlusions. 

For normal supervision, we also employ a scale-invariant loss that directly measures the alignment between rendered and predicted normals based on their cosine distance:
\begin{align}
    \mathcal{L}_{\text{normal}} = 1 - \frac{1}{HW}\sum_{i = 1}^{H}\sum_{j = 1}^{W}\frac{\hat{\mathbf{N}}_{i,j} \cdot \mathbf{N}_{i,j}}{\|\hat{\mathbf{N}}_{i,j}\| \|\mathbf{N}_{i,j}\|}.
\end{align}
$\hat{\mathbf{N}}_{i,j}$ and $\mathbf{N}_{i,j}$ represent the rendered surface normal and the pseudo GT normal at pixel $(i,j)$. The pseudo GT normal is derived from the predicted depth map $\mathbf{D}$ by applying PCA on the projected point clouds to estimate the normal direction. $H$ and $W$ are the height and width of the rendered normal image.

\mypar{Geometry-Consistent Loss}
We further enhance the reconstruction geometry consistency by introducing a novel Geometry-Consistent Loss. This loss function aims to enforce smoothness and maintain structural integrity by ensuring that the normal vectors of adjacent pixels align consistently. This is particularly important in regions with minimal depth variation where the surfaces should appear continuous. The proposed Geometry-Consistent Loss $\mathcal{L}_{\mathrm{geo}}$ is defined as:
\begin{align}
\begin{split}
    \mathcal{L}_{\text{geo}} &= \frac{\sum_{i, j} w_{i, j} \cdot \left(1 - \hat{\mathbf{N}}_{i, j} \cdot \hat{\mathbf{N}}_{i+\Delta x, j+\Delta y}\right)}{\sum_{i, j} w_{i, j}}, \\
    w_{i, j} &= 1- \left\Vert\sqrt{(\nabla_x \mathbf{D}_{i,j})^2 + (\nabla_y \mathbf{D}_{i,j})^2}\right\Vert.
\end{split}
\end{align}
Here, $(i + \Delta x, j + \Delta y)$ are the coordinates of adjacent pixels to pixel $(i,j)$, which include right and bottom neighbors ($\Delta x, \Delta y \in \{0, 1\}$). Weight $w_{i, j}$ is computed from the local depth gradients, which assign higher weights to pixels in regions with less depth change to ensure normal consistency.



Inspired by 2DGS~\cite{Huang2DGS2024}, we further regularize 3D gaussian into 2D disk shape to better represent the scene geometry by minimizing its shortest axis scale $S_i = \mathrm{diag}(s_1,s_2,s_3)$ with $\mathcal{L}_{\mathrm{scale}} = \frac{1}{N}\sum_{i \in N}\| \min(s_1, s_2, s_3) \|$, here $N$ represents the number of Gaussian splats. By combining the scale-invariant depth loss and normal loss, our method achieves robust supervision that enhances the accuracy and consistency of 3D scene reconstructions. Our final optimization loss can be defined as:
\begin{align} 
\label{eq:loss-recon}
    &\mathcal{L}_{\mathrm{total}} = \mathcal{L}_{\mathrm{rgb}} + \mathcal{L}_{\mathrm{depth}} + \mathcal{L}_{\mathrm{normal}} + \mathcal{L}_{\mathrm{geo}} + \mathcal{L}_{\mathrm{scale}}. 
\end{align}

\mypar{Screen-Space Covariance Culling}
During RL training, the agent must explore and interact with the environment to learn a robust navigation policy. However, such random explorations often lead to significant discrepancies between the agent's camera trajectory and the views encountered during training. These substantial changes in viewing angles can result in rendering artifacts and floaters, particularly near the ground surface. (Shown in Supp Sec.~\ref{supp-sec:culling})
These floaters may obstruct the agent's view and be mistakenly considered as the obstacles that block the robot and create unexpected collisions.

To mitigate these issues, we propose a simple yet effective screen-space covariance culling technique that adjusts the visual input by selectively removing artifacts based on the size of Gaussian splats when rasterized to 2D space: $\left\| \Sigma' \right\|_{\infty} > \alpha \cdot A_{\text{img}}$, where \( \left\| \Sigma' \right\|_{\infty} \) denotes the maximum norm of the covariance matrix and \( A_{\text{img}} = H \times W \) represents the total image area with the image height $H$ and width $W$. $\alpha$ is a proportionality constant that scales the threshold relative to the image size.
This approach effectively filters out splats that exceed the defined image portion, helping to maintain visual clarity and mitigate the impact of artifacts in simulation due to view discrepancy. 

\subsection{Realistic and Interactive Simulation}
An effective simulation environment should enable agents to interact with the environment and adapt to environment changes for closed-loop evaluation. The standard GS representation, however, is unsuitable as a simulator because it lacks support for agent-environment interactions.

\begin{figure*}[t]
    \centering
    \includegraphics[width=0.98\linewidth]{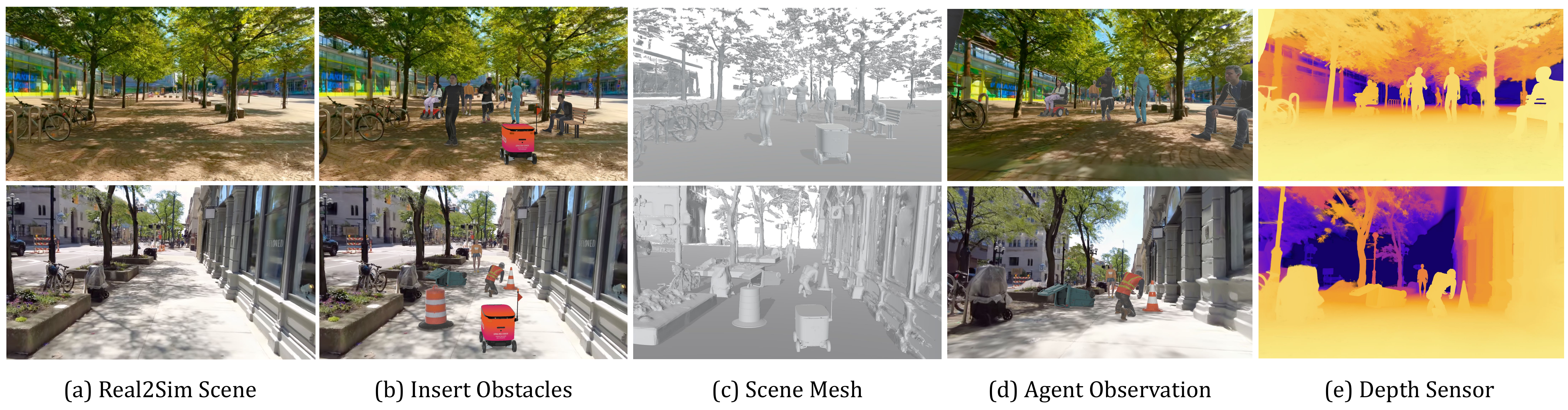}
    \vspace{-2mm}
     \caption{Interactive Scene Composition with \ourmodel: Our method is able to combine reconstructed environments with 3D assets to create diverse simulation scenarios. Here we show the (a) original real2sim environment, (b) interactive scene composition with static and dynamic obstacles, (c) scene mesh for physical collision detection, (d) agent's RGB observations, and (e) depth rendering from our hybrid scene representation that could serve as an extra sensory modality.
     }
    \label{fig:demo}
    \vspace{-2mm}
\end{figure*}

\mypar{Hybrid Scene Representation} To enhance the interactivity and realism of our simulation, as illustrated in Fig.~\ref{fig:model}, we introduce a hybrid scene representation that combines our GS representation with its scene mesh primitives to create a realistic and interactive simulation environment. As shown in Fig.~\ref{fig:demo}(c), our geometry-consistent GS reconstruction enables the extraction of high-quality environment meshes. In our hybrid scene representation, the GS provides photo-realistic visual observations for our agent training, and scene mesh is baked to support physical interaction and accurate collision detection to ensure interactive agent-environment interaction.

Specifically, we use the Truncated Signed Distance Function (TSDF)~\cite{curless1996tsdf} to export a high-quality mesh from our GS representation and employ Unity engine~\cite{haas2014history} to provide real-time physics simulation. During simulation, the GS representation and the extracted scene mesh are imported concurrently. We use a custom Unity shader to support real-time photo-realistic rendering from GS, and the mesh material is set to be invisible and operates as the collision and agent interaction primitive without impacting visual fidelity. This hybrid design combines the strength of both GS and mesh representation to support physical interaction while maintaining high-quality visual rendering for effective agent training.

\mypar{Interactive Scene Composition} While our hybrid scene representation offers realistic visuals and basic physical interactions, it does not fully capture the dynamic and complex situations that agents may encounter in real-world environments, such as new obstacles and other road users.
To closely simulate real-world navigation scenarios, we incorporate two types of obstacles into our simulation environments: 1) static objects to simulate fixed obstacles and 2) dynamic agents to emulate other road users typically present in real-world settings. 

As shown in Fig.~\ref{fig:demo}, during training, static objects such as traffic cones, trash bins, traffic lights, and poles are randomly selected and placed as obstacles within the scene. These common objects are frequently found in urban environments and greatly increase the complexity and diversity of our simulation environments with real navigation challenges. We achieve seamless composition between foreground objects and the background GS scene by combining GS rasterization with mesh rendering for both RGB and depth views. The occlusion relationships between the foreground objects and background scene are handled through z-buffering~\cite{catmull1974subdivision} to ensure depth consistency and accurate object visibility. The dynamic obstacles like pedestrians are imported and programmed with $A^*$ planning algorithm to move between random points within the scene following the shortest path, providing interactions that the agents must effectively predict and respond to. These introduced obstacles significantly increase the diversity of our environments and enable diverse interactions between the training agent and the environment, creating a comprehensive, data-rich setting that is crucial for effective navigation training.

Moreover, as shown in the second row of Fig.~\ref{fig:demo}, our framework can simulate various safety-critical scenarios, such as falling trash cans or road construction involving roadblocks and workers. These situations are often considered out-of-distribution corner cases in autonomous navigation research~\cite{bai2023on, zhou2022longtail} and are extremely crucial for training a safe and robust navigation policy. 

\begin{figure}[t]
    \centering
    \includegraphics[width=0.95\linewidth]{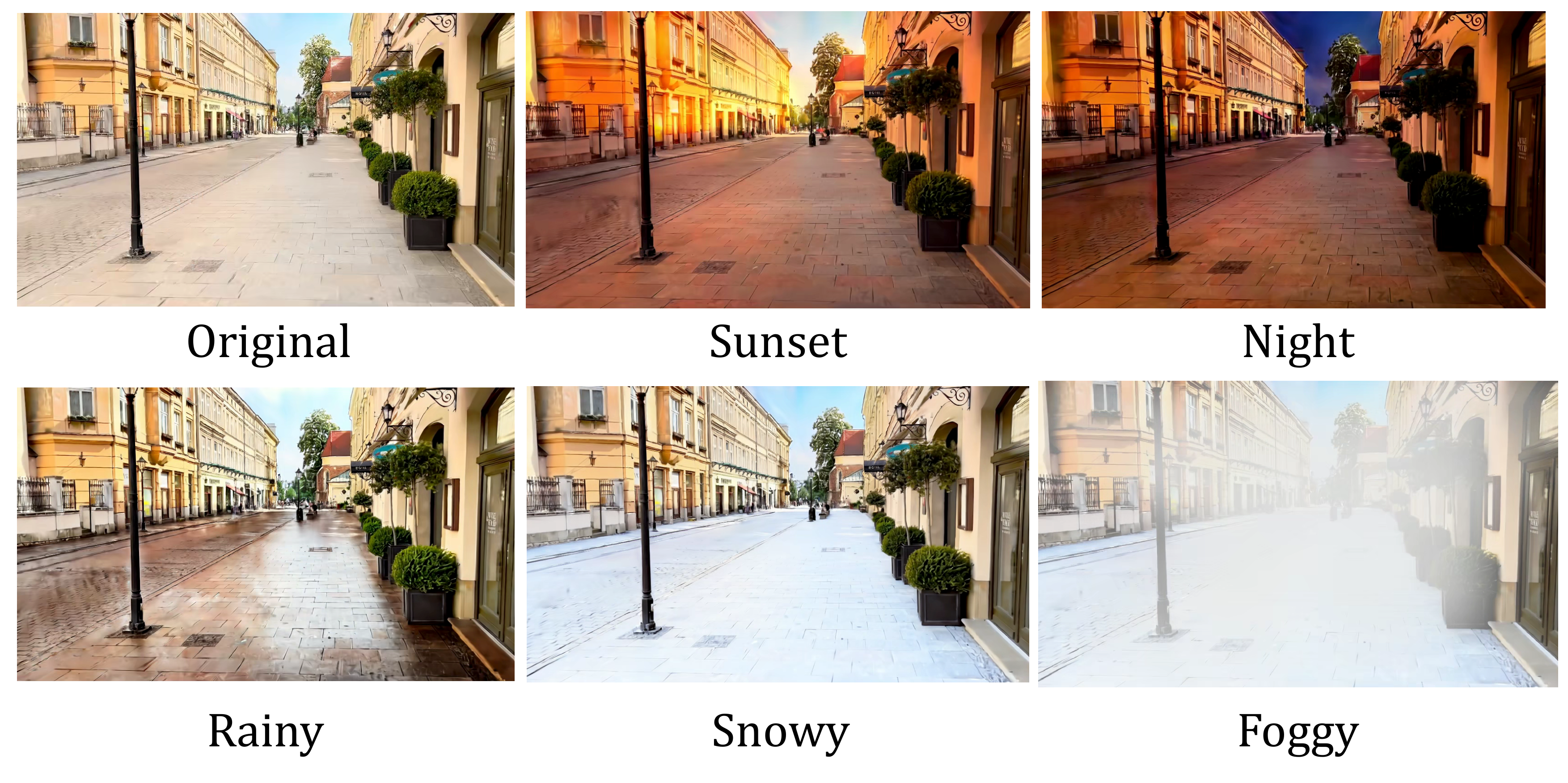}
    \vspace{-2mm}
     \caption{Scene augmentation with various overall stylization}
    \label{fig:aug}
    \vspace{-4mm}
\end{figure}

\mypar{Diverse Scene Augmentation}
Enhancing the training environment with a multi-level augmentation approach introduces greater variability and realism to agent learning. On the scene level, we integrate and extend the popular scene editing method~\cite{mou2024instruct} to support video-length consistent editing with improved temporal consistency. In this way, we can augment the training environments and generate multiple diverse scene variants that capture changes in lighting, seasons, and semantics (Fig.~\ref{fig:aug}). Additionally, similar to ClimateNeRF~\cite{Li2023ClimateNeRF}, \ourmodel can also support simulating different weather conditions such as rain, fog, and snow through particle simulation. 
With 3D layout editing and advanced weather simulations, we can create diverse, realistic environments that enrich our dataset. This variety helps our agent develop a general and robust navigation strategy across different scenarios.


\section{Experiments} 
\begin{figure*}[t]
    \centering
    \includegraphics[width=0.98\linewidth]{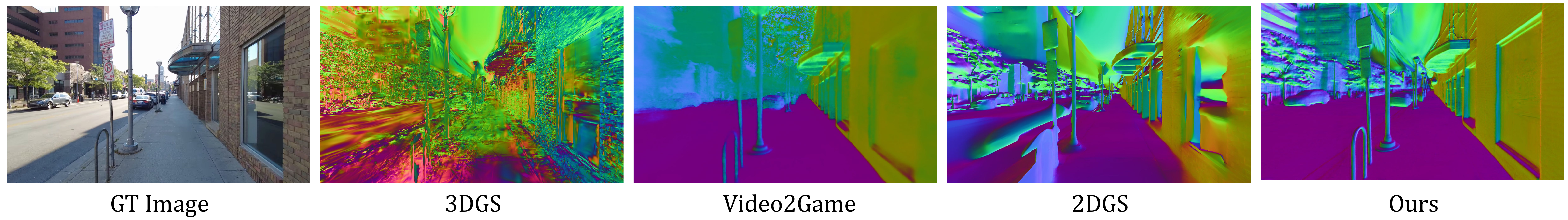}
    \vspace{-2mm}
     \caption{Surface normal renderings of different methods: results show that our approach reconstructs scene geometry with finer surface details and less artifacts compared to 3DGS~\cite{kerbl3Dgaussians}, Video2Game~\cite{xia2024video2game}, and 2DGS~\cite{Huang2DGS2024}.
     }
    \label{fig:recon-comp}
    \vspace{-3mm}
\end{figure*}
To evaluate the effectiveness of \ourmodel, we curate a dataset of real2sim environments by reconstructing 30 diverse scenes from web-sourced videos. These videos capture a range of complex urban scenarios, providing a robust foundation for in-the-wild visual navigation training. Please see the supplementary for the dataset preview. 
Our experiments are designed to evaluate three main aspects of \ourmodel's effectiveness: 
1) its ability to reconstruct high-quality 3D environments from monocular videos for simulation, 
2) its capability to support the training of robust navigation agents, and 
3) its effectiveness in minimizing the sim2real gap. We conduct extensive evaluations in both simulated and real-world environments. Results show that \ourmodel effectively builds high-quality simulations for embodied navigation training. Agents trained with our pipeline demonstrate better performance and exhibit a significantly reduced sim-to-real gap compared with agents trained in conventional environments with mesh-based observations.

\subsection{Reconstruction Evaluation}

We first evaluate our geometry-consistent reconstruction method by comparing it with other state-of-the-art reconstruction methods on our \ourmodel dataset. The \ourmodel dataset contains 30 diverse scenes. Each scene is a 15-second video at 30 fps, containing 450 frames. In every eight frames, there is one frame reserved for testing, resulting in a total of 393 training images and 57 test images per scene. As shown in Tab.~\ref{tab:recon}, our method consistently outperforms all compared methods in terms of PSNR, SSIM, and LPIPS. In Fig.~\ref{fig:recon-comp}, we further show the qualitative comparison between our method and other methods in surface normal rendering to compare the surface reconstruction quality between different methods. Our method can reconstruct finer and more accurate geometric details that align with the real-world scene compared with other methods.
\begin{table}[t]
    \centering
    \resizebox{1\linewidth}{!}{
    \begin{tabular}{l|ccc|ccc}
    \toprule
    \multirow{2}{*}{Methods}  &  \multicolumn{3}{c|}{Rendering Quality}  & \multicolumn{3}{c}{Simulation Capability} \\
    & PSNR $\uparrow$ & SSIM $\uparrow$ & LPIPS $\downarrow$  & Real Time & Interactive & RL Training  \\ 
    \midrule
    Instant-NGP~\cite{mueller2022instant}  & 27.50  & 0.827 & 0.240 & \xmark & \xmark & \xmark \\
    3DGS~\cite{kerbl3Dgaussians}           & 31.85  & 0.921 & 0.136 & \cmark & \xmark & \xmark \\
    2DGS~\cite{Huang2DGS2024}              & 30.82  & 0.915 & 0.154 & \cmark & \xmark & \xmark \\
    Video2Game~\cite{xia2024video2game}    & 28.32  & 0.834 & 0.275 & \smark & \cmark & \xmark \\
    Ours                                   & \textbf{32.41}  & \textbf{0.927} & \textbf{0.127} & \cmark & \cmark & \cmark \\
    \bottomrule
    \end{tabular}
    }
    \vspace{-2mm}
    \caption{Comparison of rendering quality and simulation capability between \ourmodel and other methods. Our method achieves the highest reconstruction quality among other methods and offers the most comprehensive simulation capabilities.}
    \label{tab:recon}
    \vspace{-3mm}
\end{table}

In supplementary we further demonstrate the effectiveness of our screen-space covariance culling method. Qualitative and quantitative results demonstrate this technique can effectively remove floater artifacts that obstruct the agent's view and significantly improve agent observation quality at extreme viewing angles.

\subsection{Urban Navigation Training}
In this section, we describe our experimental setup and evaluation metrics for training and testing our agents. We aim to test the agent's navigation capability in diverse \ourmodel environments and compare the performance with the traditional mesh-based simulator and other baseline variants. Depending on the task, the agent must learn to avoid colliding with the environment, static obstacles, and dynamic pedestrians, which reflects real-world challenges.

\mypar{Experiment Setup} We deploy \ourmodel on a four-wheeled wheeled delivery robot. The robot is equipped with a front-facing RGB camera for visual observation. For comparison, an oracle agent setup with a ground-truth depth sensor is also implemented in simulation. We stack the current image observation with the past 5 timesteps to incorporate the historical information. Visual observation and the goal observation including the distance and heading angle to the goal point are then combined as the final 
 policy observation. The agent action space includes two continuous control variables normalized in $[-1,1]$: one for adjusting wheel rotation speed and the other for modulating speed.  The simulator settings are calibrated to match real-world physical constraints to ensure a reliable sim2real transferability of the navigation policies.

\begin{table}[t]
    \centering
    \vspace{-2mm}
    \resizebox{0.98\linewidth}{!}{
        \begin{tabular}{l|c|ccc|ccc}
        \toprule
        \multirow{2}{*}{Methods} & \multirow{2}{*}{Obs} & \multicolumn{3}{c|}{\textbf{PointNav}} & \multicolumn{3}{c}{\textbf{SocialNav}} \\
        & & SR $\uparrow$ & SPL $\uparrow$ & Cost $\downarrow$ & SR $\uparrow$ & SNS $\uparrow$ & Cost $\downarrow$ \\
        \midrule
        Mesh$^{\sword}$     & RGB & 48.8\% & 0.496 & 0.34 & 43.2\% & 0.991 & 1.04 \\
        \ourmodel (Oracle) & Depth & 92.0\% & 0.937 & 0.57 & 85.6\% & 0.992 & 0.75 \\
        \midrule
        \ourmodel (No Obj) & RGB & 68.8\% & 0.695 & 1.45 & 61.6\% & 0.973 & 1.79 \\
        \ourmodel (Static) & RGB & 80.8\% & 0.818 & 0.94 & 71.2\% & 0.980 & 1.74 \\
        \ourmodel (Dynamic) & RGB & \textbf{81.6}\% & \textbf{0.824} & \textbf{0.86} & \textbf{74.4}\% & \textbf{0.987} & \textbf{1.21} \\
        \bottomrule
        \end{tabular}
    }
    \vspace{-2mm}
    \caption{Evaluation of agent navigation performance in simulation. Mesh$^{\sword}$ represents only use our reconstructed mesh for visual observation to simulate the mesh-based simulation method~\cite{xia2024video2game}.}
    \label{tab:main}
    \vspace{-3mm}
\end{table}

\mypar{Tasks and metrics} 
We design two common navigation tasks, Point Navigation (\textit{PointNav}) and Social Navigation (\textit{SocialNav}), to test agent performance in both static and dynamic scenarios. In both tasks, the agent must navigate through the environment from a starting location to a goal point which are randomized across different episodes to ensure robust policy learning. For \textit{PointNav}, the agent needs to avoid hitting the environment and static obstacles placed within the scene, while in \textit{SocialNav}, the agent must avoid colliding with both static obstacles and other moving pedestrians by adapting its path according to their movements.

Our agents are trained across 30 unique \ourmodel simulation environments with the off-policy reinforcement learning algorithm Soft-Actor-Critic (SAC)~\cite{haarnoja2017soft} for 1.5M steps and tested on 5 hold-out testing scenes which the agent has never seen during training. 
The agent's performance on each task is evaluated based on the success rate (SR)~\cite{anderson2018sr}, success rate weighted by path length (SPL)~\cite{batra2020spl}, hit number (Cost) and social navigation score (SNS)~\cite{deitke2022retrospectivesembodiedaiworkshop}. All trials are rolled out 25 times to mitigate result variance. For each trial, 0 to 5 static objects are randomly sampled and placed between the agent starting point and goal point as obstacles. For social navigation dynamic pedestrians are placed within the scene and animated to walk across the movable area randomly. We report the average score across all 125 trials. Specific details including hyper-parameter settings, reward shaping, training, and testing scene descriptions are available in supplementary Sec.~\ref{supp-sec:rl-training}.

\begin{figure}[t]
    \centering
    \includegraphics[width=0.98\linewidth]{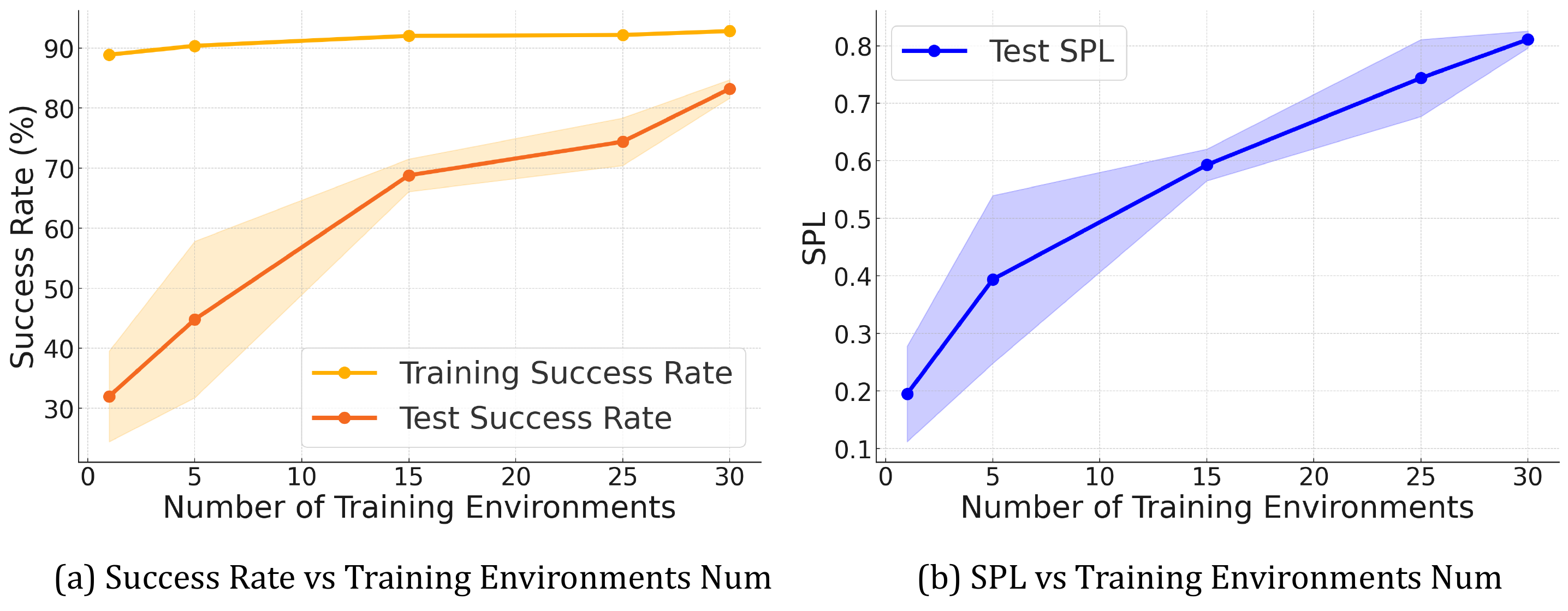}
    \vspace{-2mm}
     \caption{Generalization Results: This table compares (a) the success rate (SR) and (b) success rate weighted by path length (SPL) across varying numbers of training environments. Increasing the number of training environments leads to a higher test success rate and SPL, which indicates improved agents generalizability.
     }
     \label{fig:generalization}
     \vspace{-5mm}
\end{figure}

\mypar{Performance Evaluation}
As there is no existing method that can convert videos into a functional simulation environment for RL training, we simulate a comparable mesh-based approach using our reconstructed colored mesh representation for agent visual navigation, similar to the approach taken by Video2Game~\cite{xia2024video2game} for game development.

We report the performance of agents trained under three conditions: without obstacles, with static obstacles, and with both static and dynamic obstacles, and compare their results across these setups. Additionally, we evaluate agents trained with the oracle depth observations to better understand the capabilities of different policies. Tab.~\ref{tab:main} shows that the agents trained with both static and dynamic obstacles achieved the best performance with an 81.6\% success rate on the \textit{PointNav} task and a 74.4\% on the \textit{SocialNav} task. It achieves a significant 32.8\% and 31.2\% performance improvement compared to traditional mesh-based simulation, respectively. Notably, even when trained only with static obstacles, our agent still demonstrates emergent capabilities in \textit{SocialNav} task by reaching a 71.2\% success rate.

Fig.~\ref{fig:generalization} shows that our agent generalizability is substantially improved by increasing the number of training environments, where we evaluate performance across 1, 5, 15, 25, and 30 training environments. The test scene success rate increases significantly when training with 15 environments and continues to rise with additional training environments. For SPL, the trend is similar. In general, the variances of SR and SPL continue to decrease as the number of training environments increases, indicating more robust navigation performance.

\subsection{Sim-to-Real Deployment}
\label{sec:exp-sim2real}
\begin{figure}[t]
    \centering
    \includegraphics[width=0.9\linewidth]{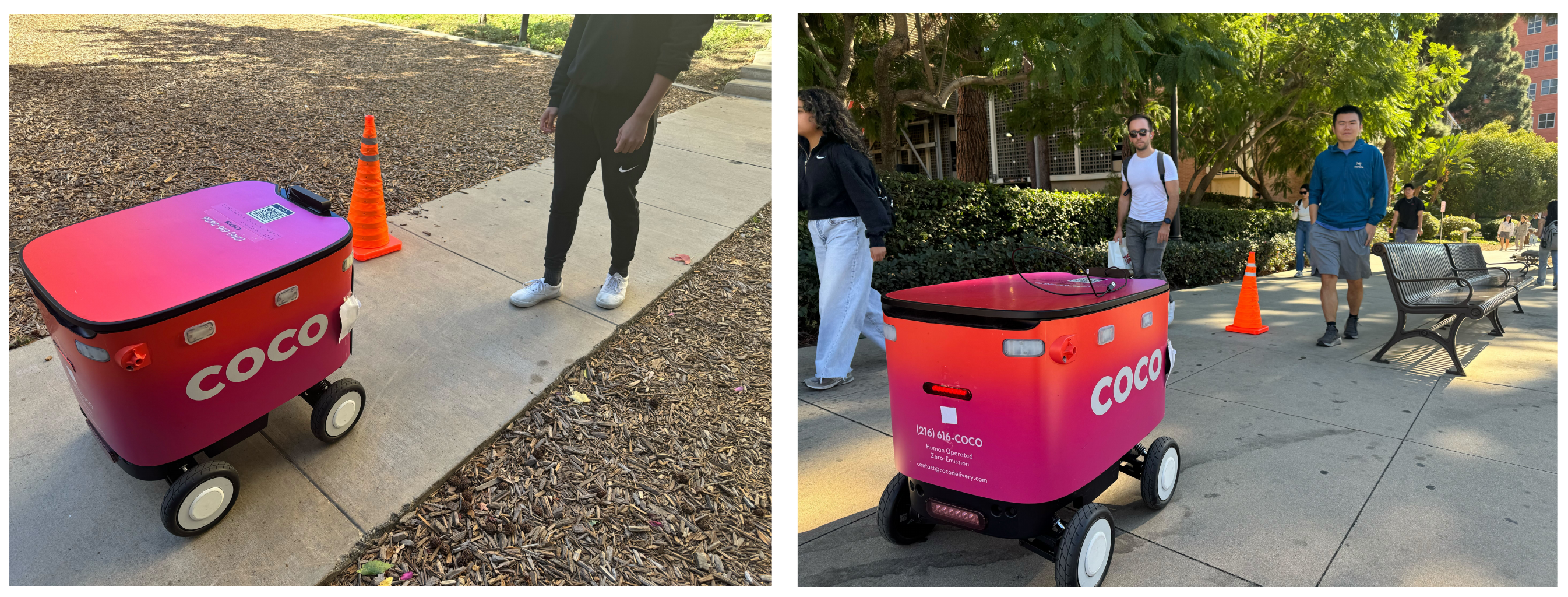}
    \vspace{-2mm}
     \caption{Real-world experiment settings}
    \label{fig:realworld}
    \vspace{-1mm}
\end{figure}

To evaluate the effectiveness of the proposed \ourmodel pipeline in bridging the sim-to-real gap, we deploy agents trained in \ourmodel environments to the real world in zero-shot settings. Fig.~\ref{fig:realworld} demonstrates our real-world experiment settings in diverse environments.
We further show a qualitative comparison between the simulation and real-world digital-twin environments in supplementary Fig.~\ref{supp-fig:sim-real-comp} to illustrate our reduced sim-to-real gap. Other real-world experiment details can be found in supplementary Sec.~\ref{supp-sec:sim2real-details}

\begin{table}[t]
    \centering
    \vspace{-2mm}
     \resizebox{0.95\linewidth}{!}{

    \begin{tabular}{l|ccc}
    \toprule
    \text{Method (Env N)} & Go Straight & Static Obstacle & Dynamic Obstacle
    \\ \midrule
    Baseline (30)        &  0\%  & 0\%  &  0\%  \\
    Vid2Sim (1)        &  0\%  & 30\%  & 0\%   \\
    Vid2Sim (5)        &  60\% & 40\%  &  0\%  \\
   Vid2Sim (30)       &  \textbf{85\%} & \textbf{65\%} & \textbf{55\%}  \\
    \bottomrule
    \end{tabular}
    }
    \vspace{-2mm}
    \caption{The navigation performance of trained agents in the real world. The number in the parenthesis indicates the number of environments the agent is trained on.}
    \label{tab:exp_real_world_1}
    \vspace{-4mm}

\end{table}

\mypar{Experiment Setup} We test the policy in real-world urban environments which is not included in the \ourmodel training dataset. Each environment has a walkable region of at least 3m in width and 20m in length, sufficient for the robot to avoid obstacles. In each trial, we randomly sample starting and ending positions in the walkable region with a travel distance between 5m and 15m and evaluate the performance on the following tasks: 1) \textit{Go Straight}: reach the goal in an empty environment without obstacles. 2) \textit{Static Obstacle}: reach the goal while avoiding static objects placed in between the starting and goal points. 3). \textit{Dynamic Obstacle}: reach the goal while avoiding dynamic pedestrians walking towards the robot. The robot succeeds if it reaches within 0.5m of the goal and fails if it either collides with other objects or moves out of the walkable region. We perform 20 trials for each task and report the final success rate.
\mypar{Results} As shown in Tab.~\ref{tab:exp_real_world_1}, the mesh reconstruction baseline fails to complete all three tasks. This shows the large sim-to-real gap with the RGB observation from mesh representation. On the contrary, Vid2Sim performs better by training with more environments, similar to what is observed in Fig~\ref{fig:generalization}. Note that all results are from zero-shot deployment without fine-tuning. This highlights the effectiveness of Vid2Sim in addressing the sim-to-real gap and the possibility of learning a general real-world deployable visual navigation policy in simulation by adopting the Vid2Sim pipeline to more videos.

\section{Conclusion}
This work introduces a novel framework \ourmodel that can convert monocular videos into a realistic and interactive simulation environment for learning robust and generalizable urban navigation policies. Experiments show our method significantly reduces the sim-to-real gap and provides a scalable solution to create simulation environments from real-world scenes. We hope to extend this framework to many other agents like legged robots in the future.

{
    \small
    \bibliographystyle{ieeenat_fullname}
    \bibliography{main}

\begin{thebibliography}{71}
\providecommand{\natexlab}[1]{#1}
\providecommand{\url}[1]{\texttt{#1}}
\expandafter\ifx\csname urlstyle\endcsname\relax
  \providecommand{\doi}[1]{doi: #1}\else
  \providecommand{\doi}{doi: \begingroup \urlstyle{rm}\Url}\fi

\bibitem[Anderson et~al.(2018)Anderson, Chang, Chaplot, Dosovitskiy, Gupta, Koltun, Kosecka, Malik, Mottaghi, Savva, and Zamir]{anderson2018sr}
Peter Anderson, Angel Chang, Devendra~Singh Chaplot, Alexey Dosovitskiy, Saurabh Gupta, Vladlen Koltun, Jana Kosecka, Jitendra Malik, Roozbeh Mottaghi, Manolis Savva, and Amir~R. Zamir.
\newblock On evaluation of embodied navigation agents.
\newblock \emph{arXiv preprint arXiv:1807.06757}, 2018.

\bibitem[Bai et~al.(2023)Bai, Liu, Wang, Hao, FENG, Chu, and Hu]{bai2023on}
Jianhong Bai, Zuozhu Liu, Hualiang Wang, Jin Hao, YANG FENG, Huanpeng Chu, and Haoji Hu.
\newblock On the effectiveness of out-of-distribution data in self-supervised long-tail learning.
\newblock In \emph{ICLR}, 2023.

\bibitem[Barron et~al.(2022)Barron, Mildenhall, Verbin, Srinivasan, and Hedman]{barron2022mipnerf360}
Jonathan~T. Barron, Ben Mildenhall, Dor Verbin, Pratul~P. Srinivasan, and Peter Hedman.
\newblock Mip-nerf 360: Unbounded anti-aliased neural radiance fields.
\newblock In \emph{CVPR}, 2022.

\bibitem[Batra et~al.(2020)Batra, Gokaslan, Kembhavi, Maksymets, Mottaghi, Savva, Toshev, and Wijmans]{batra2020spl}
Dhruv Batra, Aaron Gokaslan, Aniruddha Kembhavi, Oleksandr Maksymets, Roozbeh Mottaghi, Manolis Savva, Alexander Toshev, and Erik Wijmans.
\newblock Objectnav revisited: On evaluation of embodied agents navigating to objects.
\newblock \emph{arXiv preprint arXiv:2006.13171}, 2020.

\bibitem[Birkl et~al.(2023)Birkl, Wofk, and M{\"u}ller]{birkl2023midas}
Reiner Birkl, Diana Wofk, and Matthias M{\"u}ller.
\newblock Midas v3.1 -- a model zoo for robust monocular relative depth estimation.
\newblock \emph{PAMI}, 2023.

\bibitem[Brockman et~al.(2016)Brockman, Cheung, Pettersson, Schneider, Schulman, Tang, and Zaremba]{brockman2016openai}
Greg Brockman, Vicki Cheung, Ludwig Pettersson, Jonas Schneider, John Schulman, Jie Tang, and Wojciech Zaremba.
\newblock Openai gym.
\newblock \emph{arXiv preprint arXiv:1606.01540}, 2016.

\bibitem[Catmull(1974)]{catmull1974subdivision}
Edwin Catmull.
\newblock \emph{A Subdivision Algorithm for Computer Display of Curved Surfaces}.
\newblock Ph.d. dissertation, University of Utah, Salt Lake City, UT, 1974.

\bibitem[Chebotar et~al.(2019)Chebotar, Handa, Makoviychuk, Macklin, Issac, Ratliff, and Fox]{Chebotar2019Closing}
Yevgen Chebotar, Ankur Handa, Viktor Makoviychuk, Miles Macklin, Jan Issac, Nathan Ratliff, and Dieter Fox.
\newblock Closing the sim-to-real loop: Adapting simulation randomization with real world experience.
\newblock In \emph{ICRA}, 2019.

\bibitem[Cheng et~al.(2023)Cheng, Oh, Price, Schwing, and Lee]{cheng2023tracking}
Ho~Kei Cheng, Seoung~Wug Oh, Brian Price, Alexander Schwing, and Joon-Young Lee.
\newblock Tracking anything with decoupled video segmentation.
\newblock In \emph{ICCV}, 2023.

\bibitem[Chung et~al.(2023)Chung, Oh, and Lee]{chung2023depth}
Jaeyoung Chung, Jeongtaek Oh, and Kyoung~Mu Lee.
\newblock Depth-regularized optimization for 3d gaussian splatting in few-shot images.
\newblock In \emph{CVPRW}, 2023.

\bibitem[Curless and Levoy(1996)]{curless1996tsdf}
Brian Curless and Marc Levoy.
\newblock A volumetric method for building complex models from range images.
\newblock In \emph{Proceedings of the 23rd Annual Conference on Computer Graphics and Interactive Techniques}, 1996.

\bibitem[Deitke et~al.(2022)Deitke, Batra, Bisk, Campari, Chang, Chaplot, Chen, D'Arpino, Ehsani, Farhadi, Fei-Fei, Francis, Gan, Grauman, Hall, Han, Jain, Kembhavi, Krantz, Lee, Li, Majumder, Maksymets, Martín-Martín, Mottaghi, Raychaudhuri, Roberts, Savarese, Savva, Shridhar, Sünderhauf, Szot, Talbot, Tenenbaum, Thomason, Toshev, Truong, Weihs, and Wu]{deitke2022retrospectivesembodiedaiworkshop}
Matt Deitke, Dhruv Batra, Yonatan Bisk, Tommaso Campari, Angel~X. Chang, Devendra~Singh Chaplot, Changan Chen, Claudia~Pérez D'Arpino, Kiana Ehsani, Ali Farhadi, Li Fei-Fei, Anthony Francis, Chuang Gan, Kristen Grauman, David Hall, Winson Han, Unnat Jain, Aniruddha Kembhavi, Jacob Krantz, Stefan Lee, Chengshu Li, Sagnik Majumder, Oleksandr Maksymets, Roberto Martín-Martín, Roozbeh Mottaghi, Sonia Raychaudhuri, Mike Roberts, Silvio Savarese, Manolis Savva, Mohit Shridhar, Niko Sünderhauf, Andrew Szot, Ben Talbot, Joshua~B. Tenenbaum, Jesse Thomason, Alexander Toshev, Joanne Truong, Luca Weihs, and Jiajun Wu.
\newblock Retrospectives on the embodied ai workshop.
\newblock \emph{arXiv preprint arXiv:2210.06849}, 2022.

\bibitem[Dosovitskiy et~al.(2017)Dosovitskiy, Ros, Codevilla, Lopez, and Koltun]{alexey2017carla}
Alexey Dosovitskiy, German Ros, Felipe Codevilla, Antonio Lopez, and Vladlen Koltun.
\newblock Carla: An open urban driving simulator.
\newblock In \emph{CoRL}, 2017.

\bibitem[Gu{\'e}don and Lepetit(2024)]{guedon2023sugar}
Antoine Gu{\'e}don and Vincent Lepetit.
\newblock Sugar: Surface-aligned gaussian splatting for efficient 3d mesh reconstruction and high-quality mesh rendering.
\newblock In \emph{CVPR}, 2024.

\bibitem[Haarnoja et~al.(2017)Haarnoja, Zhou, Abbeel, and Levine]{haarnoja2017soft}
Tuomas Haarnoja, Aurick Zhou, Pieter Abbeel, and Sergey Levine.
\newblock Soft actor-critic: Off-policy maximum entropy deep reinforcement learning with a stochastic actor.
\newblock \emph{Deep Reinforcement Learning Symposium}, 2017.

\bibitem[Haas(2014)]{haas2014history}
John~K Haas.
\newblock A history of the unity game engine.
\newblock \emph{Worcester Polytechnic Institute}, 2014.

\bibitem[Heusel et~al.(2017)Heusel, Ramsauer, Unterthiner, Nessler, and Hochreiter]{heusel2018ganstrainedtimescaleupdate}
Martin Heusel, Hubert Ramsauer, Thomas Unterthiner, Bernhard Nessler, and Sepp Hochreiter.
\newblock Gans trained by a two time-scale update rule converge to a local nash equilibrium.
\newblock In \emph{NeurIPS}, 2017.

\bibitem[Ho et~al.(2020)Ho, Jain, and Abbeel]{ho2020ddpm}
Jonathan Ho, Ajay Jain, and Pieter Abbeel.
\newblock Denoising diffusion probabilistic models.
\newblock In \emph{NeurIPS}, 2020.

\bibitem[Huang et~al.(2024)Huang, Yu, Chen, Geiger, and Gao]{Huang2DGS2024}
Binbin Huang, Zehao Yu, Anpei Chen, Andreas Geiger, and Shenghua Gao.
\newblock 2d gaussian splatting for geometrically accurate radiance fields.
\newblock In \emph{SIGGRAPH}, 2024.

\bibitem[Izadi et~al.(2011)Izadi, Kim, Hilliges, Molyneaux, Newcombe, Kohli, Shotton, Hodges, Freeman, Davison, and Fitzgibbon]{shahram2011kinectfusion}
Shahram Izadi, David Kim, Otmar Hilliges, David Molyneaux, Richard Newcombe, Pushmeet Kohli, Jamie Shotton, Steve Hodges, Dustin Freeman, Andrew Davison, and Andrew Fitzgibbon.
\newblock Kinectfusion: real-time 3d reconstruction and interaction using a moving depth camera.
\newblock In \emph{Proceedings of the 24th Annual ACM Symposium on User Interface Software and Technology}, 2011.

\bibitem[Jain et~al.(2021)Jain, Tancik, and Abbeel]{Jain_2021_ICCV}
Ajay Jain, Matthew Tancik, and Pieter Abbeel.
\newblock Putting nerf on a diet: Semantically consistent few-shot view synthesis.
\newblock In \emph{ICCV}, 2021.

\bibitem[Kalashnikov et~al.(2018)Kalashnikov, Irpan, Pastor, Ibarz, Herzog, Jang, Quillen, Holly, Kalakrishnan, Vanhoucke, and Levine]{pmlr-v87-kalashnikov18a}
Dmitry Kalashnikov, Alex Irpan, Peter Pastor, Julian Ibarz, Alexander Herzog, Eric Jang, Deirdre Quillen, Ethan Holly, Mrinal Kalakrishnan, Vincent Vanhoucke, and Sergey Levine.
\newblock Scalable deep reinforcement learning for vision-based robotic manipulation.
\newblock In \emph{Proceedings of The 2nd Conference on Robot Learning}. PMLR, 2018.

\bibitem[Ke et~al.(2023)Ke, Ye, Danelljan, Liu, Tai, Tang, and Yu]{sam_hq}
Lei Ke, Mingqiao Ye, Martin Danelljan, Yifan Liu, Yu-Wing Tai, Chi-Keung Tang, and Fisher Yu.
\newblock Segment anything in high quality.
\newblock In \emph{NeurIPS}, 2023.

\bibitem[Kerbl et~al.(2023)Kerbl, Kopanas, Leimk{\"u}hler, and Drettakis]{kerbl3Dgaussians}
Bernhard Kerbl, Georgios Kopanas, Thomas Leimk{\"u}hler, and George Drettakis.
\newblock 3d gaussian splatting for real-time radiance field rendering.
\newblock \emph{ACM Transactions on Graphics}, 2023.

\bibitem[Kirillov et~al.(2023)Kirillov, Mintun, Ravi, Mao, Rolland, Gustafson, Xiao, Whitehead, Berg, Lo, Doll{\'a}r, and Girshick]{kirillov2023sam}
Alexander Kirillov, Eric Mintun, Nikhila Ravi, Hanzi Mao, Chloe Rolland, Laura Gustafson, Tete Xiao, Spencer Whitehead, Alexander~C. Berg, Wan-Yen Lo, Piotr Doll{\'a}r, and Ross Girshick.
\newblock Segment anything.
\newblock \emph{arXiv:2304.02643}, 2023.

\bibitem[Li et~al.(2022)Li, Peng, Feng, Zhang, Xue, and Zhou]{li2021metadrive}
Quanyi Li, Zhenghao Peng, Lan Feng, Qihang Zhang, Zhenghai Xue, and Bolei Zhou.
\newblock Metadrive: Composing diverse driving scenarios for generalizable reinforcement learning.
\newblock \emph{IEEE Transactions on Pattern Analysis and Machine Intelligence}, 2022.

\bibitem[Li et~al.(2023{\natexlab{a}})Li, Zhang, and Ye]{li2023drivingdiffusion}
Xiaofan Li, Yifu Zhang, and Xiaoqing Ye.
\newblock Drivingdiffusion: Layout-guided multi-view driving scene video generation with latent diffusion model.
\newblock \emph{arXiv preprint arXiv:2310.07771}, 2023{\natexlab{a}}.

\bibitem[Li et~al.(2024)Li, Hsu, Gu, Pertsch, Mees, Walke, Fu, Lunawat, Sieh, Kirmani, Levine, Wu, Finn, Su, Vuong, and Xiao]{li24simpler}
Xuanlin Li, Kyle Hsu, Jiayuan Gu, Karl Pertsch, Oier Mees, Homer~Rich Walke, Chuyuan Fu, Ishikaa Lunawat, Isabel Sieh, Sean Kirmani, Sergey Levine, Jiajun Wu, Chelsea Finn, Hao Su, Quan Vuong, and Ted Xiao.
\newblock Evaluating real-world robot manipulation policies in simulation.
\newblock \emph{arXiv preprint arXiv:2405.05941}, 2024.

\bibitem[Li et~al.(2023{\natexlab{b}})Li, Lin, Forsyth, Huang, and Wang]{Li2023ClimateNeRF}
Yuan Li, Zhi-Hao Lin, David Forsyth, Jia-Bin Huang, and Shenlong Wang.
\newblock Climatenerf: Extreme weather synthesis in neural radiance field.
\newblock In \emph{ICCV}, 2023{\natexlab{b}}.

\bibitem[Li et~al.(2023{\natexlab{c}})Li, M\"uller, Evans, Taylor, Unberath, Liu, and Lin]{li2023neuralangelo}
Zhaoshuo Li, Thomas M\"uller, Alex Evans, Russell~H Taylor, Mathias Unberath, Ming-Yu Liu, and Chen-Hsuan Lin.
\newblock Neuralangelo: High-fidelity neural surface reconstruction.
\newblock In \emph{CVPR}, 2023{\natexlab{c}}.

\bibitem[Ma et~al.(2023)Ma, Meng, Liu, Chen, Xu, and Chen]{liqian2023sim2real2}
Liqian Ma, Jiaojiao Meng, Shuntao Liu, Weihang Chen, Jing Xu, and Rui Chen.
\newblock Sim2real2: Actively building explicit physics model for precise articulated object manipulation.
\newblock In \emph{ICRA}, 2023.

\bibitem[Makoviychuk et~al.(2021)Makoviychuk, Wawrzyniak, Guo, Lu, Storey, Macklin, Hoeller, Rudin, Allshire, Handa, and State]{makoviychuk2021isaacgymhighperformance}
Viktor Makoviychuk, Lukasz Wawrzyniak, Yunrong Guo, Michelle Lu, Kier Storey, Miles Macklin, David Hoeller, Nikita Rudin, Arthur Allshire, Ankur Handa, and Gavriel State.
\newblock Isaac gym: High performance gpu-based physics simulation for robot learning.
\newblock \emph{arXiv preprint arXiv:2108.10470}, 2021.

\bibitem[Martin-Brualla et~al.(2021)Martin-Brualla, Radwan, Sajjadi, Barron, Dosovitskiy, and Duckworth]{martinbrualla2020nerfw}
Ricardo Martin-Brualla, Noha Radwan, Mehdi S.~M. Sajjadi, Jonathan~T. Barron, Alexey Dosovitskiy, and Daniel Duckworth.
\newblock {NeRF in the Wild: Neural Radiance Fields for Unconstrained Photo Collections}.
\newblock In \emph{CVPR}, 2021.

\bibitem[Miangoleh et~al.(2024)Miangoleh, Reddy, and Aksoy]{miangolehSIDepth}
S.~Mahdi~H. Miangoleh, Mahesh Reddy, and Ya\u{g}{\i}z Aksoy.
\newblock Scale-invariant monocular depth estimation via ssi depth.
\newblock In \emph{Proc. SIGGRAPH}, 2024.

\bibitem[Mildenhall et~al.(2020)Mildenhall, Srinivasan, Tancik, Barron, Ramamoorthi, and Ng]{mildenhall2020nerf}
Ben Mildenhall, Pratul~P. Srinivasan, Matthew Tancik, Jonathan~T. Barron, Ravi Ramamoorthi, and Ren Ng.
\newblock Nerf: Representing scenes as neural radiance fields for view synthesis.
\newblock In \emph{ECCV}, 2020.

\bibitem[Mou et~al.(2024)Mou, Chen, and Wang]{mou2024instruct}
Linzhan Mou, Jun-Kun Chen, and Yu-Xiong Wang.
\newblock Instruct 4d-to-4d: Editing 4d scenes as pseudo-3d scenes using 2d diffusion.
\newblock In \emph{CVPR}, 2024.

\bibitem[M\"uller et~al.(2022)M\"uller, Evans, Schied, and Keller]{mueller2022instant}
Thomas M\"uller, Alex Evans, Christoph Schied, and Alexander Keller.
\newblock Instant neural graphics primitives with a multiresolution hash encoding.
\newblock \emph{ACM Transactions on Graphics}, 2022.

\bibitem[Niemeyer et~al.(2022)Niemeyer, Barron, Mildenhall, Sajjadi, Geiger, and Radwan]{Niemeyer2021Regnerf}
Michael Niemeyer, Jonathan~T. Barron, Ben Mildenhall, Mehdi S.~M. Sajjadi, Andreas Geiger, and Noha Radwan.
\newblock Regnerf: Regularizing neural radiance fields for view synthesis from sparse inputs.
\newblock In \emph{CVPR}, 2022.

\bibitem[Pan et~al.(2024)Pan, Barath, Pollefeys, and Sch\"{o}nberger]{pan2024glomap}
Linfei Pan, Daniel Barath, Marc Pollefeys, and Johannes~Lutz Sch\"{o}nberger.
\newblock {Global Structure-from-Motion Revisited}.
\newblock In \emph{ECCV}, 2024.

\bibitem[Peng et~al.(2018)Peng, Andrychowicz, Zaremba, and Abbeel]{Peng2018SimToReal}
Xue~Bin Peng, Marcin Andrychowicz, Wojciech Zaremba, and Pieter Abbeel.
\newblock Sim-to-real transfer of robotic control with dynamics randomization.
\newblock In \emph{ICRA}, 2018.

\bibitem[Raffin et~al.(2021)Raffin, Hill, Gleave, Kanervisto, Ernestus, and Dormann]{stable-baselines3}
Antonin Raffin, Ashley Hill, Adam Gleave, Anssi Kanervisto, Maximilian Ernestus, and Noah Dormann.
\newblock Stable-baselines3: Reliable reinforcement learning implementations.
\newblock \emph{Journal of Machine Learning Research}, 2021.

\bibitem[Sadeghi and Levine(2017)]{Sadeghi2017CAD2RL}
Fereshteh Sadeghi and Sergey Levine.
\newblock {CAD2RL}: Real single-image flight without a single real image.
\newblock In \emph{RSS}, 2017.

\bibitem[Savva et~al.(2019)Savva, Kadian, Maksymets, Zhao, Wijmans, Jain, Straub, Liu, Koltun, Malik, Parikh, and Batra]{habitat19iccv}
Manolis Savva, Abhishek Kadian, Oleksandr Maksymets, Yili Zhao, Erik Wijmans, Bhavana Jain, Julian Straub, Jia Liu, Vladlen Koltun, Jitendra Malik, Devi Parikh, and Dhruv Batra.
\newblock Habitat: {A} {P}latform for {E}mbodied {AI} {R}esearch.
\newblock In \emph{ICCV}, 2019.

\bibitem[Sch\"{o}nberger and Frahm(2016)]{schoenberger2016sfm}
Johannes~Lutz Sch\"{o}nberger and Jan-Michael Frahm.
\newblock Structure-from-motion revisited.
\newblock In \emph{Conference on Computer Vision and Pattern Recognition (CVPR)}, 2016.

\bibitem[Shen et~al.(2023)Shen, Chandaka, Lin, Zhai, Cui, Forsyth, and Wang]{shen2023simonwheels}
Yuan Shen, Bhargav Chandaka, Zhi-Hao Lin, Albert Zhai, Hang Cui, David Forsyth, and Shenlong Wang.
\newblock Sim-on-wheels: Physical world in the loop simulation for self-driving.
\newblock In \emph{IEEE Robotics and Automation Letters}, 2023.

\bibitem[Song et~al.(2021)Song, Meng, and Ermon]{song2021ddim}
Jiaming Song, Chenlin Meng, and Stefano Ermon.
\newblock Denoising diffusion implicit models.
\newblock In \emph{ICLR}, 2021.

\bibitem[Swerdlow et~al.(2024)Swerdlow, Xu, and Zhou]{swerdlow2024streetview}
Alexander Swerdlow, Runsheng Xu, and Bolei Zhou.
\newblock Street-view image generation from a bird's-eye view layout.
\newblock \emph{IEEE Robotics and Automation Letters}, 2024.

\bibitem[Szot et~al.(2021)Szot, Clegg, Undersander, Wijmans, Zhao, Turner, Maestre, Mukadam, Chaplot, Maksymets, Gokaslan, Vondrus, Dharur, Meier, Galuba, Chang, Kira, Koltun, Malik, Savva, and Batra]{szot2021habitat}
Andrew Szot, Alex Clegg, Eric Undersander, Erik Wijmans, Yili Zhao, John Turner, Noah Maestre, Mustafa Mukadam, Devendra Chaplot, Oleksandr Maksymets, Aaron Gokaslan, Vladimir Vondrus, Sameer Dharur, Franziska Meier, Wojciech Galuba, Angel Chang, Zsolt Kira, Vladlen Koltun, Jitendra Malik, Manolis Savva, and Dhruv Batra.
\newblock Habitat 2.0: Training home assistants to rearrange their habitat.
\newblock In \emph{NeurIPS}, 2021.

\bibitem[Tan et~al.(2018)Tan, Zhang, Coumans, Iscen, Bai, Hafner, Bohez, and Vanhoucke]{Tan2018SimToReal}
Ji Tan, Tingnan Zhang, Erwin Coumans, Atil Iscen, Yunfei Bai, Danijar Hafner, Steven Bohez, and Vincent Vanhoucke.
\newblock Sim-to-real: Learning agile locomotion for quadruped robots.
\newblock In \emph{RSS}, 2018.

\bibitem[Tobin et~al.(2017)Tobin, Fong, Ray, Schneider, Zaremba, and Abbeel]{domainRandom2017}
Josh Tobin, Rachel Fong, Alex Ray, Jonas Schneider, Wojciech Zaremba, and Pieter Abbeel.
\newblock Domain randomization for transferring deep neural networks from simulation to the real world.
\newblock In \emph{IROS}, 2017.

\bibitem[Todorov et~al.(2012)Todorov, Erez, and Tassa]{todorov2012mujoco}
Emanuel Todorov, Tom Erez, and Yuval Tassa.
\newblock Mujoco: A physics engine for model-based control.
\newblock In \emph{2012 IEEE/RSJ International Conference on Intelligent Robots and Systems}, 2012.

\bibitem[Tremblay et~al.(2018)Tremblay, Prakash, Acuna, Brophy, Jampani, Huynh, To, Cameracci, Boochoon, and Birchfield]{Tremblay2018Training}
Jonathan Tremblay, Aayush Prakash, David Acuna, Mark Brophy, Varun Jampani, Cong~Phuoc Huynh, Timo To, Eric Cameracci, Steve Boochoon, and Stan Birchfield.
\newblock Training deep networks with synthetic data: Bridging the reality gap by domain randomization.
\newblock In \emph{CVPRW}, 2018.

\bibitem[Wang et~al.(2021)Wang, Liu, Liu, Theobalt, Komura, and Wang]{wang2021neus}
Peng Wang, Lingjie Liu, Yuan Liu, Christian Theobalt, Taku Komura, and Wenping Wang.
\newblock Neus: Learning neural implicit surfaces by volume rendering for multi-view reconstruction.
\newblock In \emph{NeurIPS}, 2021.

\bibitem[Wang et~al.(2024)Wang, Zhu, Huang, Chen, Zhu, and Lu]{wang2023drive}
Xiaofeng Wang, Zheng Zhu, Guan Huang, Xinze Chen, Jiagang Zhu, and Jiwen Lu.
\newblock Drivedreamer: Towards real-world-driven world models for autonomous driving.
\newblock In \emph{ECCV}, 2024.

\bibitem[Wolf et~al.(2024)Wolf, Bracha, and Kimmel]{wolf2024gs2mesh}
Yaniv Wolf, Amit Bracha, and Ron Kimmel.
\newblock Gs2mesh: Surface reconstruction from gaussian splatting via novel stereo views.
\newblock In \emph{ECCV}, 2024.

\bibitem[Wu et~al.(2024{\natexlab{a}})Wu, Zheng, and Cai]{Wu2024gsrec}
Qianyi Wu, Jianmin Zheng, and Jianfei Cai.
\newblock Surface reconstruction from 3d gaussian splatting via local structural hints.
\newblock In \emph{ECCV}, 2024{\natexlab{a}}.

\bibitem[Wu et~al.(2024{\natexlab{b}})Wu, He, Wang, Duan, He, Liu, Li, and Zhou]{wu2024metaurban}
Wayne Wu, Honglin He, Yiran Wang, Chenda Duan, Jack He, Zhizheng Liu, Quanyi Li, and Bolei Zhou.
\newblock Metaurban: A simulation platform for embodied ai in urban spaces.
\newblock \emph{arXiv preprint arXiv:2407.08725}, 2024{\natexlab{b}}.

\bibitem[Xia et~al.(2018)Xia, R.~Zamir, He, Sax, Malik, and Savarese]{xiazamirhe2018gibsonenv}
Fei Xia, Amir R.~Zamir, Zhi-Yang He, Alexander Sax, Jitendra Malik, and Silvio Savarese.
\newblock Gibson {Env}: real-world perception for embodied agents.
\newblock In \emph{CVPR}, 2018.

\bibitem[Xia et~al.(2024)Xia, Lin, Ma, and Wang]{xia2024video2game}
Hongchi Xia, Zhi-Hao Lin, Wei-Chiu Ma, and Shenlong Wang.
\newblock Video2game: Real-time, interactive, realistic and browser-compatible environment from a single video.
\newblock In \emph{CVPR}, 2024.

\bibitem[Xu et~al.(2024)Xu, Yi, Xu, Tao, Ong, and Zhang]{xu2024fewshotnerfadaptiverendering}
Qingshan Xu, Xuanyu Yi, Jianyao Xu, Wenbing Tao, Yew-Soon Ong, and Hanwang Zhang.
\newblock Few-shot nerf by adaptive rendering loss regularization.
\newblock In \emph{ECCV}, 2024.

\bibitem[Yang et~al.(2023{\natexlab{a}})Yang, Pavone, and Wang]{Yang2023FreeNeRF}
Jiawei Yang, Marco Pavone, and Yue Wang.
\newblock Freenerf: Improving few-shot neural rendering with free frequency regularization.
\newblock In \emph{CVPR}, 2023{\natexlab{a}}.

\bibitem[Yang et~al.(2024{\natexlab{a}})Yang, Kang, Huang, Xu, Feng, and Zhao]{depth_anything_v1}
Lihe Yang, Bingyi Kang, Zilong Huang, Xiaogang Xu, Jiashi Feng, and Hengshuang Zhao.
\newblock Depth anything: Unleashing the power of large-scale unlabeled data.
\newblock In \emph{CVPR}, 2024{\natexlab{a}}.

\bibitem[Yang et~al.(2024{\natexlab{b}})Yang, Kang, Huang, Zhao, Xu, Feng, and Zhao]{depth_anything_v2}
Lihe Yang, Bingyi Kang, Zilong Huang, Zhen Zhao, Xiaogang Xu, Jiashi Feng, and Hengshuang Zhao.
\newblock Depth anything v2.
\newblock \emph{arXiv preprint arXiv:2406.09414}, 2024{\natexlab{b}}.

\bibitem[Yang et~al.(2023{\natexlab{b}})Yang, Chen, Wang, Manivasagam, Ma, Yang, and Urtasun]{yang2023unisim}
Ze Yang, Yun Chen, Jingkang Wang, Sivabalan Manivasagam, Wei-Chiu Ma, Anqi~Joyce Yang, and Raquel Urtasun.
\newblock Unisim: A neural closed-loop sensor simulator.
\newblock In \emph{CVPR}, 2023{\natexlab{b}}.

\bibitem[Ye et~al.(2024)Ye, Li, Liu, Qiao, and Dou]{ye2024absgs}
Zongxin Ye, Wenyu Li, Sidun Liu, Peng Qiao, and Yong Dou.
\newblock Absgs: Recovering fine details for 3d gaussian splatting.
\newblock \emph{arXiv preprint arXiv:2404.10484}, 2024.

\bibitem[Yu et~al.(2024{\natexlab{a}})Yu, Yang, Choi, Ravan, Leonard, and Isola]{yu2024lucidsim}
Alan Yu, Ge Yang, Ran Choi, Yajvan Ravan, John Leonard, and Phillip Isola.
\newblock Lucidsim: Learning agile visual locomotion from generated images.
\newblock In \emph{CORL}, 2024{\natexlab{a}}.

\bibitem[Yu et~al.(2024{\natexlab{b}})Yu, Sattler, and Geiger]{Yu2024GOF}
Zehao Yu, Torsten Sattler, and Andreas Geiger.
\newblock Gaussian opacity fields: Efficient adaptive surface reconstruction in unbounded scenes.
\newblock \emph{ACM Transactions on Graphics}, 2024{\natexlab{b}}.

\bibitem[Zhang et~al.(2024)Zhang, Wang, Wang, Li, Qin, and Wang]{zhang2024gaussian}
Dongbin Zhang, Chuming Wang, Weitao Wang, Peihao Li, Minghan Qin, and Haoqian Wang.
\newblock Gaussian in the wild: 3d gaussian splatting for unconstrained image collections.
\newblock In \emph{ECCV}, 2024.

\bibitem[Zhou et~al.(2022)Zhou, Cao, Xu, Deng, Liu, Jiang, and Yang]{zhou2022longtail}
Weitao Zhou, Zhong Cao, Yunkang Xu, Nanshan Deng, Xiaoyu Liu, Kun Jiang, and Diange Yang.
\newblock Long-tail prediction uncertainty aware trajectory planning for self-driving vehicles.
\newblock In \emph{2022 IEEE 25th International Conference on Intelligent Transportation Systems (ITSC)}, 2022.

\bibitem[Zhou et~al.(2024)Zhou, Simon, Peng, Mo, Zhu, Guo, and Zhou]{zhou2024simgen}
Yunsong Zhou, Michael Simon, Zhenghao Peng, Sicheng Mo, Hongzi Zhu, Minyi Guo, and Bolei Zhou.
\newblock Simgen: Simulator-conditioned driving scene generation.
\newblock In \emph{NeurIPS}, 2024.

\bibitem[Zhu et~al.(2024)Zhu, Fan, Jiang, and Wang]{zhu2023FSGS}
Zehao Zhu, Zhiwen Fan, Yifan Jiang, and Zhangyang Wang.
\newblock Fsgs: Real-time few-shot view synthesis using gaussian splatting.
\newblock In \emph{ECCV}, 2024.

\end{thebibliography}
}

\clearpage
\setcounter{page}{1}
\setcounter{section}{0}
\maketitleappendix

\def\thesection{\Alph{section}}

\section{\ourmodel dataset}
\label{supp-sec:dataset}

\begin{figure*}[t]
    \centering
    \includegraphics[width=0.98\linewidth]{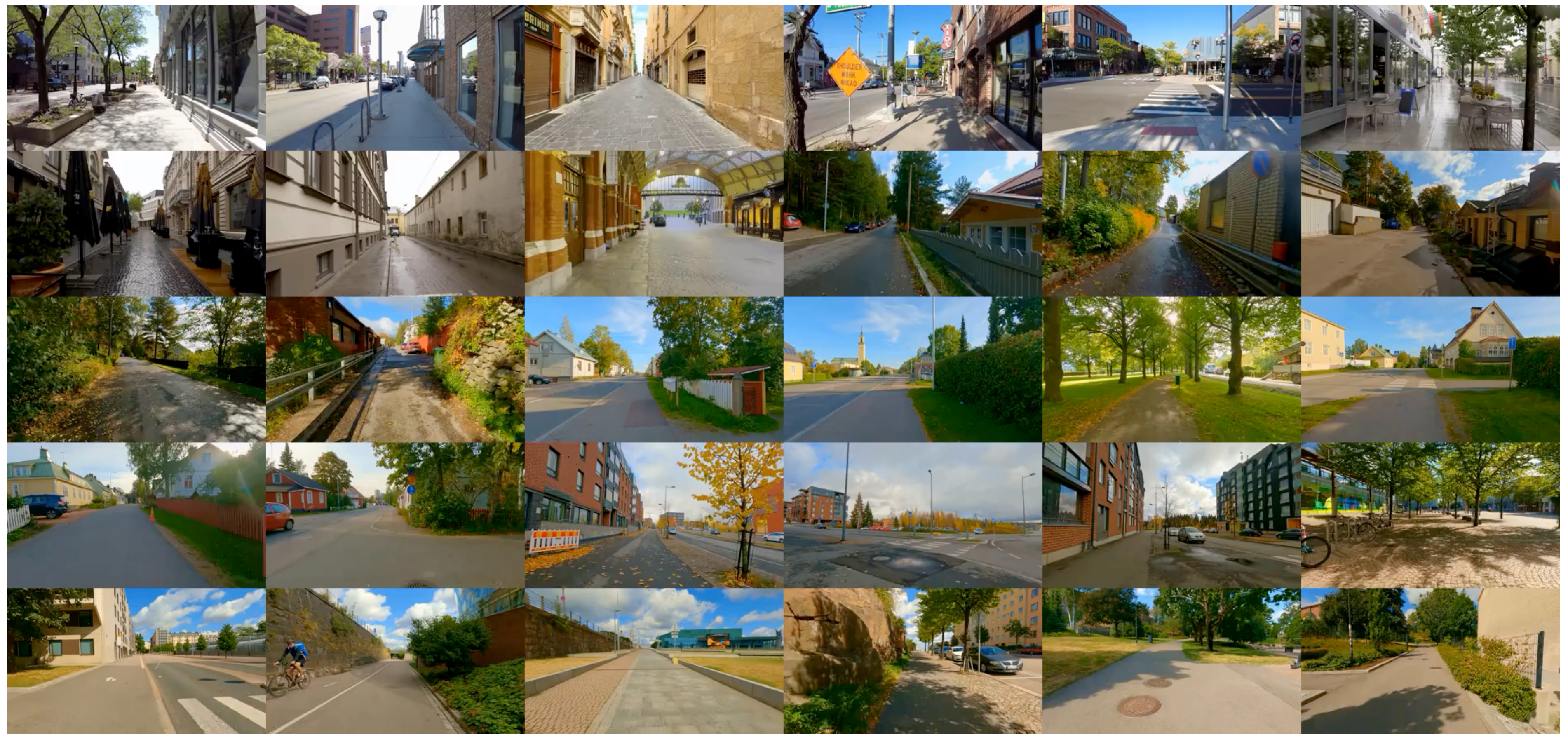}
     \vspace{-2mm}
     \caption{\ourmodel Dataset Preview}
     \label{supp-fig:dataset}
     \vspace{-2mm}
\end{figure*}

Our \ourmodel dataset includes 30 high-quality real-to-sim simulation environments. These environments are reconstructed from video clips sourced from 9 web videos recorded by individuals walking along streets with a single hand-held camera. Each clip capture includes 15 seconds of forward-facing video recorded at 30 fps, providing 450 frames per scene for reconstruction.
To ensure privacy, we mask all human faces and identifiable information in the video frames. Commonly used Structure-from-Motion (SfM) methods, such as COLMAP~\cite{schoenberger2016sfm}, often fail to reconstruct accurate camera poses from in-the-wild videos due to significant variations in viewpoints and lighting conditions. Instead, we employ GLOMAP~\cite{pan2024glomap}, an advanced general-purpose SfM system that is more robust and efficient than COLMAP, to obtain accurate camera poses.

In Fig.~\ref{supp-fig:dataset}, We show a preview snap-shot of the 30 diverse environments in our dataset, these environments encompass a variety of urban navigation scenarios designed for more robust and generalizable navigation policy training in complex urban environments.

\section{Geometry-Consistent Reconstruction}
\label{supp-sec:recon}

\subsection{Implementation details}
In this section, we mainly introduce the implementation details of our geometry-consistent reconstruction method, including training strategy, hyper-parameter settings, and other extra details. 

To avoid the influence of dynamic objects within the original videos, for each frame, we generate a dynamic mask to mask out all the moving objects within the scene with an off-the-shelf video-based tracker DEVA~\cite{cheng2023tracking}. We train our model for 30000 iterations. We use the same training hyper-parameters as 3DGS~\cite{kerbl3Dgaussians} and adopt the densification strategy from AbsGS~\cite{ye2024absgs}. All the depth, normal, and geometry-consistency loss are started to apply at the 500 iterations still the end of the training. 

For the mesh extraction, we first render the depth map from each frame and fuse them into a TSDF field with KinectFusion~\cite{shahram2011kinectfusion}, the mesh is then extracted from TSDF field with voxel size set to 0.1. To eliminate the potential dynamics issue caused by uneven ground reconstructions, we further remove the ground plane of the scene in our mesh extraction through ground plane segmentation. We found that directly applying a general segmentation model (such as SAM~\cite{kirillov2023sam} or SAM-HQ~\cite{sam_hq}) cannot remove all the ground mesh due to inaccurate segmentation and prompt ambiguity. Instead, we propose to generate a detailed ground mask $\mathcal{M}_{i}$ by using the ground plane's normal direction as a prior:
\begin{align}
    \mathcal{M}_{i} = \|\arccos(\mathbf{N}_i\cdot \Bar{\mathbf{n}}_{i}^{'}) < \delta\|,
\end{align}
where $\mathbf{N}_i$ indicates the rendered normal map of the $i^{th}$ frame, $\Bar{\mathbf{n}}_{i}^{'}$ denotes the mean normal direction of the ground surface computed from the ground segmentation mask given by the SAM-HQ~\cite{sam_hq} model and $\delta$ denotes the angle threshold. Pixels whose normal directions are within $\delta$ degrees of the mean ground surface normal direction are grouped together to generate the final ground mask.Incorporating normal priors results in a more precise ground mask, enabling clean mesh extraction without the ground surface. We set $\delta = 15$ degrees in our implementation.

\subsection{Screen-space covariance culling}
\label{supp-sec:culling}

In this section, we systematically evaluate our screen-space covariance culling method both qualitatively and quantitatively. Results demonstrate this technique could effectively remove rendering artifacts and significantly improve agent observation quality.

Since floater artifacts often appear when the viewing angles and focal length differ significantly from the training view. Therefore, standard test views that closely align with the training view may not accurately represent real situations during agent training. For better evaluation, we simulate the agent's camera view by adjusting the test view's camera focal down by 1.5x and shifting the camera down by 1 unit. Given the fact that there is no ground-truth image available to apply common NVS evaluation metrics (e.g. PSNR and SSIM) to evaluate the quality of the culled images, we instead provide a quantitative evaluation of screen-space covariance culling by computing the Fréchet Inception Distance (FID)~\cite{heusel2018ganstrainedtimescaleupdate} score between the training views and the agent's rendered observations. The qualitative results show a substantial  10.7\% improvement in the FID score after our screen-space covariance culling, which supports its effectiveness.

It is important to note that, due to the limited size of the evaluation dataset and the significant difference between the agent's views and the training views, the absolute FID values cannot be directly compared to those commonly reported in generative models evaluation~\cite{ho2020ddpm, song2021ddim}. Instead, the FID score in this context serves as a relative metric for us to better understand the improvements achieved through this covariance culling process.

\begin{figure}[t]
    \centering
    \includegraphics[width=0.98\linewidth]{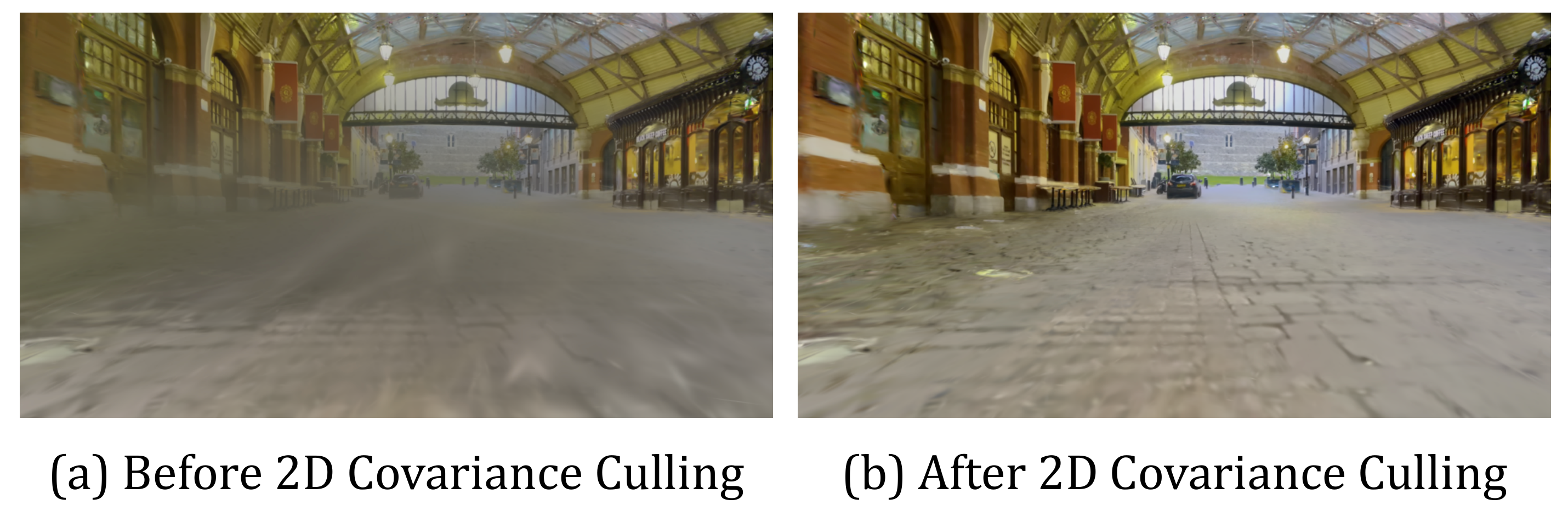}
     \vspace{-2mm}
     \caption{Screen-space Covariance Culling Comparison}
     \label{supp-fig:cull}
     \vspace{-4mm}
\end{figure}

\begin{table}[t]
    \centering
    \resizebox{0.9\linewidth}{!}{
    \begin{tabular}{l|c|c|c}
    \toprule
    Metric & w/o Culling & w Culling & Improvement (\%) \\ 
    \midrule
    FID~\cite{heusel2018ganstrainedtimescaleupdate} & 214.47 & 191.54 & +10.70\% \\
    \bottomrule
    \end{tabular}
    }
    \caption{Comparison of FID scores for rendered images with and without 2D covariance culling.}
    \label{tab:cull}
    \vspace{-2mm}
\end{table}

\section{Simulation Environment Setup}
\label{supp-sec:sim-env}

Our simulation environment is built based on the Unity~\cite{haas2014history} physics engine. Our agents, hybrid scene representation, and static and dynamic obstacles are imported together to form our diverse simulation environments. The ground surface is configured as a horizontal walkable area, while other scene meshes and obstacles are tagged as collidable objects. To maintain visual consistency, the ground and scene meshes are rendered invisible and serve only for collision interactions. 

For the agent settings, we position the agent's camera sensors at the front of the robot, matching the real-world sensor placement. The robot size in the simulation is scaled to metric scale with the background scene adjusted proportionally. The agent operates using a bicycle dynamics model, configured with a wheelbase of 0.8 meters and a maximum turning angle of 30 degrees. 
We introduce random perturbations to the camera's position and rotation to simulate real-world disturbances caused by uneven ground during navigation.

\begin{figure}[H]
    \centering
    \includegraphics[width=0.98\linewidth]{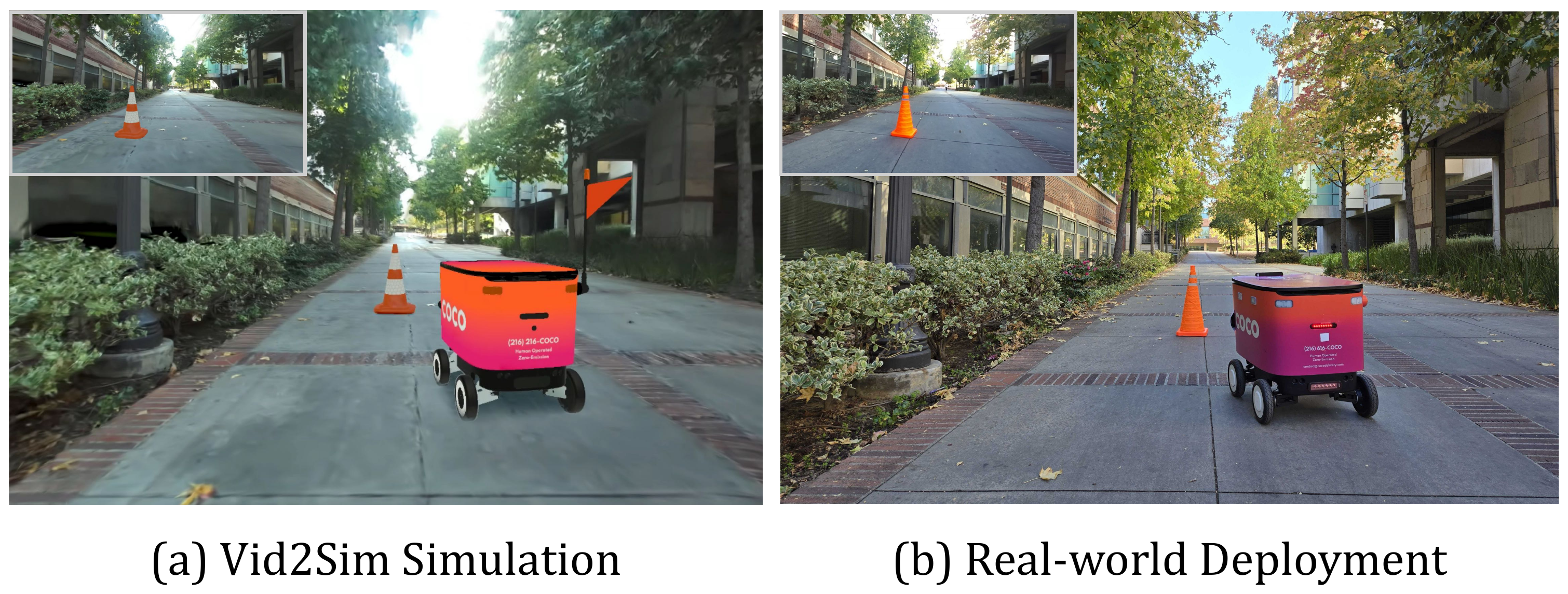}
     \vspace{-2mm}
     \caption{Digital-twin environment for real-world scene}
     \label{supp-fig:sim-real-comp}
     \vspace{-2mm}
\end{figure}

In Fig.~\ref{supp-fig:sim-real-comp}, We also provide an example of a digital twin environment created using our \ourmodel pipeline that transforms a real-world environment into a simulation environment with the minimal sim-to-real gap. The agent RGB observation is provided at the top-left corner which replicates real-world observation.

\section{RL Training Details}
\label{supp-sec:rl-training}
In this section, we provide the training details of our RL agents. To enable agent training with popular RL frameworks like OpenAI Gym~\cite{brockman2016openai} and Stable-Baselines3~\cite{stable-baselines3}, we compiled our Unity environments and integrated them with an environment wrapper to make them compatible with the OpenAI Gym interface. The simulation system is running at 50hz and the RL training is running at 5hz. 

During each episode, the agents are randomly initialized from a starting point and need to navigate to another random goal point within the scene. Typically, the distance between the starting point and the goal point is around 10$\sim$30m. For both the PointNav and SocialNav tasks, an agent is considered successful if it reaches the goal within 0.5m. 

\subsection{Reward shaping}
Our reward function for the agent is defined as follows:
\begin{align*}
    R = R_{term} + c_1R_{dist} + c_2R_{steer} + c_3R_{crash} + c_4R_{time}
\end{align*}
\begin{itemize}
    \item Terminal reward $R_{term}$: is a sparse reward set to $+10$ if the agent successfully reaches the destination and $-10$ if it fails.
    \item Distance reward $R_{dist}$: is a dense reward defined as $R_{dist} = d_t - d_{t-1}$, where $d_t$ represents the current distance between the agent to the goal point, which guide agent navigates towards the goal during training. We set the weight $c_1=1$.
    \item Steering smoothness reward $R_{steer}$: is a regularization reward defined as $R_{steer}=-\|s_t-s_{t-1}\| \cdot v_t$ to penalize inconsistent steer movement during agent navigation. The weight $c_2$ is set to $0.05$
    \item Crash reward $R_{crash}$ is used to penalize the collision between the agent and other objects, including the environment, static obstacles, and dynamic agents. It's a dense negative reward defined as $-1(c_t)$ where $c_t$ denotes the collision happens at time $t$ and $1(\cdot)$ is the indicator function. We set the weight of $R_{crash}$ as $c_3=1.0$.
    \item Time reward $R_{time}$ is a dense reward defined as $R_{time} = -\Delta t$, between two time steps the $R_{time}$ is simplifies to $-1$ to encourage efficient navigation by penalizing longer episodes. We set the weight $c_4$ of $R_{time}$ to $-0.1$
\end{itemize}

\noindent At any time step $t$ if the $R_{term} \neq 0$, the episode will terminate. The agent is considered as failed and receives a $R_{term} = -10$ under the following conditions: 1) Navigating outside the drivable area. 2) Exceeding 3000 time steps in the episode, resulting in a timeout. 3) Accumulating more than 3 collisions within an episode. During the testing, any collision between the agent and other objects will result in a cost $+1$.

\subsection{Hyper-parameter settings}
We train our agent with 30 parallel environments and the training takes around 15 hours on a single NVIDIA A5000 GPU. We provide our hyper-parameter settings in Tab.~\ref{tab:sac_hyperparameters}

\begin{table}[h]
\centering
\begin{tabular}{ll}
\hline
\textbf{SAC Hyper-parameters}      & \textbf{Value} \\ \hline
Learning starts                    & 10000          \\
$\tau$ (Target critic update ratio) & 0.005         \\
Discount factor $\gamma$           & 0.99           \\
SDE sample frequency               & 64             \\
Batch size                         & 256            \\
Learning rate                      & $3 \times 10^{-4}$ \\
Use SDE at warmup                  & True           \\
Use SDE                            & True           \\ \hline
\end{tabular}
\caption{SAC Hyper-parameters used in experiments.}
\label{tab:sac_hyperparameters}
\end{table}

\section{Sim-to-real Deployment}
\label{supp-sec:sim2real-details}
For the real-world experiment, we deploy navigation policies trained in \ourmodel on the same four-wheeled delivery robot used in the simulation. The robot takes RGB images as inputs directly from an onboard camera, which has the same resolution of 1280$\times$720 and the same intrinsic and extrinsic parameters as the sensor specifications in simulation. We resize the current image to 128$\times$72 and stack it with the images from the past 5 timesteps to incorporate the historical information. We then combine the image inputs and the distance and heading angle to the goal point from robot odometry as the policy observation. The action output includes the normalized linear and angular velocities between -1 and 1, and we perform system identification to align the dynamics of the real robot with the simulation by re-scaling the normalized velocities in real-world units, and we use the built-in controller of the robot to convert the velocity commands to the low-level motor controls for actions.

\section{Particle Simulation for Weather Editing}

\begin{figure}[t]
    \centering
    \includegraphics[width=0.98\linewidth]{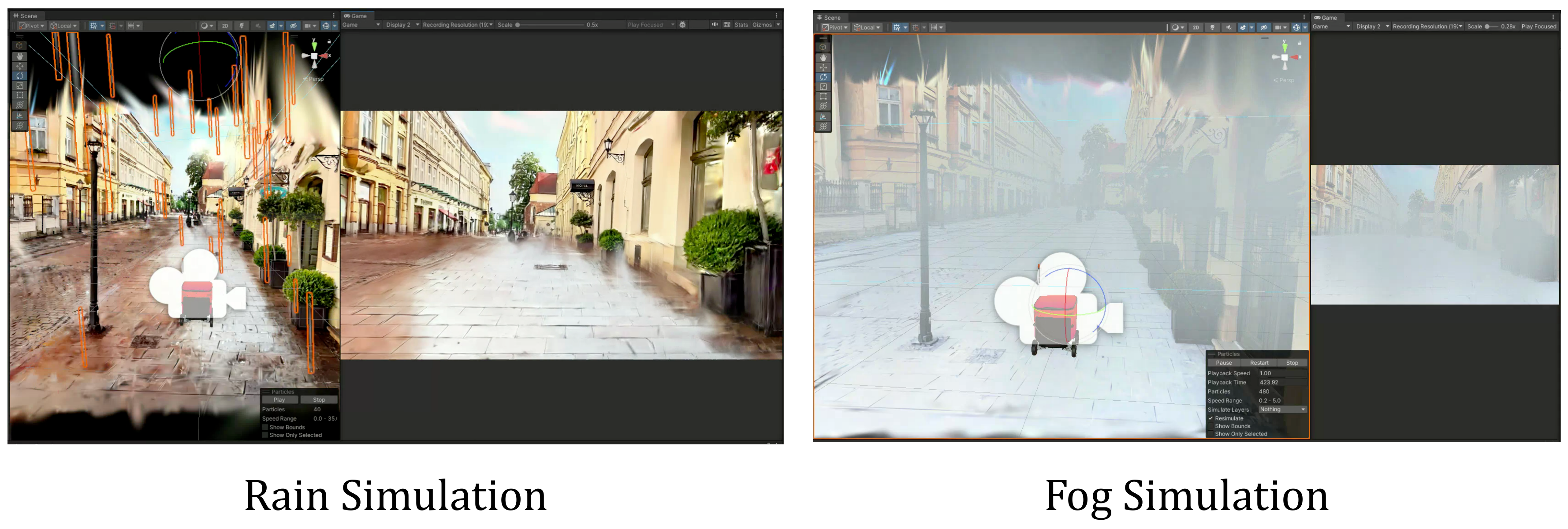}
     \vspace{-2mm}
     \caption{Weather simulation with particle systems in Unity}
     \label{supp-fig:weather_sim}
     \vspace{-4mm}
\end{figure}

Apart from global scene layout editing illustrated in the main paper, we also demonstrate our ability to simulate different weather conditions like rainy and foggy weather through 3D particle simulations within the Unity environments in Fig.~\ref{supp-fig:weather_sim}

\section{Limitations and Future works}
\label{supp-sec:limitations}

Though the proposed Vid2Sim framework can support efficient training in simulation environments with fast GS-based rendering and can achieve zero-shot sim2real deployment, building each simulation environment is time-consuming as GS requires GLOMAP~\cite{pan2024glomap} to initialize the point cloud and the camera poses, which takes a long time to run. Therefore, we have only collected 30 environments and experiment results have shown training with more environments can lead to better performance. 
In the future, we will explore more efficient ways to convert the monocular videos to GS-based simulation environments and build a larger real2sim dataset with more diverse environments.  We believe such large-scale environments can further benefit the training of a generalizable navigation policy and extend our pipeline to train other embodiments like humanoids and robot dogs.

\begin{figure*}[t]
    \centering
    \includegraphics[width=0.98\linewidth]{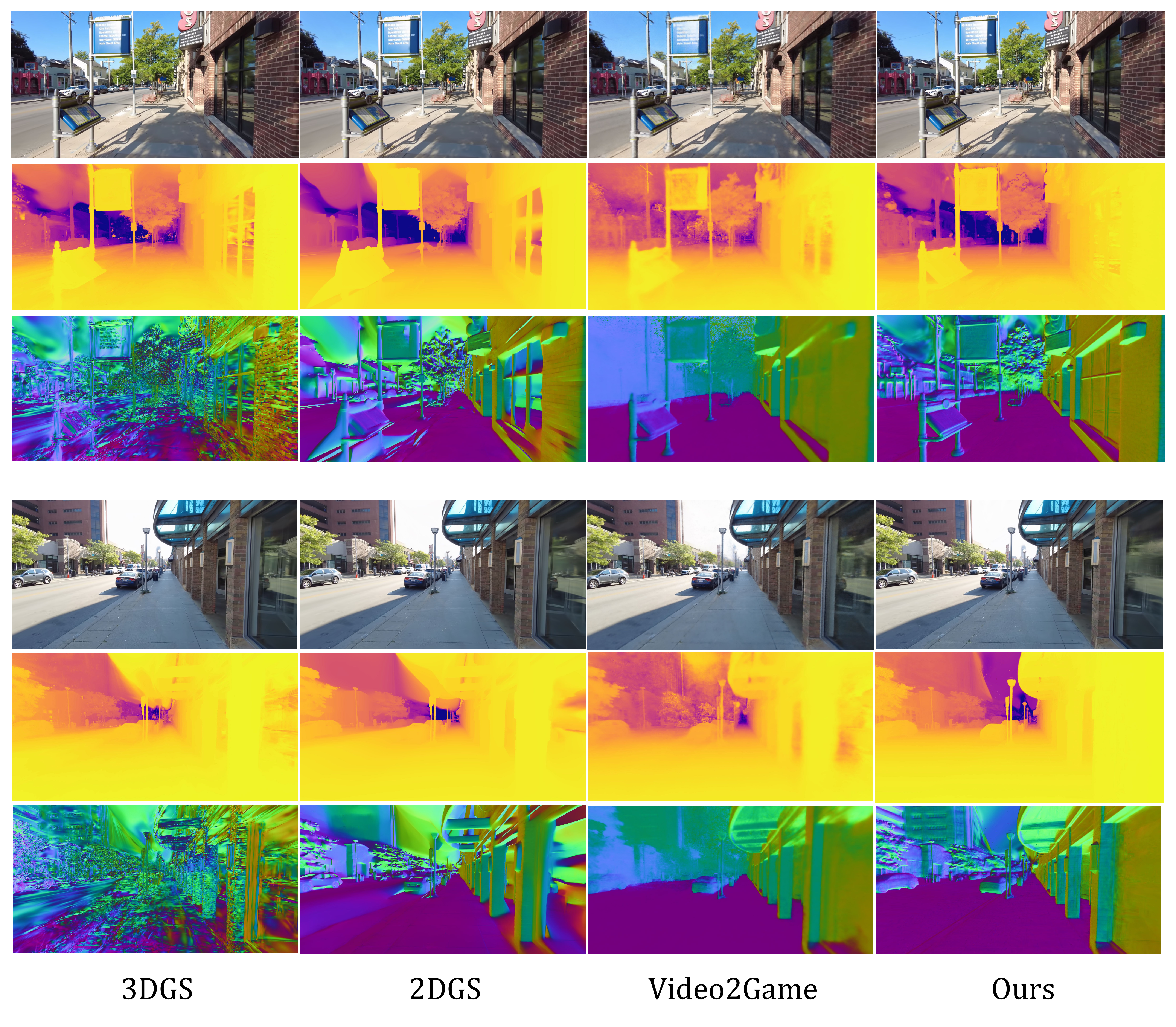}
     \vspace{-2mm}
     \caption{Reconstruction Comparison between different methods on RGB, depth map and normal map}
     \vspace{-4mm}
\end{figure*}

\clearpage


\end{document}